\documentclass[journal]{IEEEtran}
\usepackage{cite}
\ifCLASSINFOpdf
   \usepackage[pdftex]{graphicx}
\else
\fi
\usepackage{amsmath}

\usepackage{amsmath,epsfig,amsfonts,amssymb,multirow}
\usepackage[pagebackref=true,breaklinks=true,colorlinks,bookmarks=false,citecolor=blue]{hyperref}
\usepackage{booktabs}
\usepackage{graphicx}
\usepackage{enumitem}
\usepackage{bbding}
\usepackage{xcolor}
\usepackage{overpic}
\makeatletter
\newcommand{\removelatexerror}{\let\@latex@error\@gobble}
\makeatother
\usepackage{amssymb}
\usepackage{amsmath,amsfonts}
\usepackage{amsopn,bm}

\usepackage{nicefrac}       

\usepackage{epsfig}
\usepackage{graphicx}
\usepackage{float}
\usepackage{wrapfig}
\usepackage[ruled,vlined]{algorithm2e}

\usepackage{bm,xspace}
\usepackage{comment}
\usepackage{verbatim}
\usepackage{multirow}
\usepackage{makecell}
\usepackage{array}
\usepackage{balance}
\usepackage{url}
\usepackage{booktabs}
\usepackage{etoolbox,siunitx}
\usepackage{calc}
\usepackage{pifont,hologo}
 
\usepackage{nicefrac}

\usepackage{arydshln}
\usepackage[utf8]{inputenc} 
\usepackage[T1]{fontenc}    
\usepackage{url}            
\usepackage{amsfonts}       
\usepackage{nicefrac}       
\usepackage{microtype}      
\usepackage[american]{babel}
\usepackage{enumitem}
\usepackage{siunitx}

\usepackage{enumitem}

\usepackage[super]{nth}
\usepackage{nicefrac}
\robustify\bfseries

\usepackage{dsfont}

\usepackage{listings}
\usepackage{xcolor}
\definecolor{codegreen}{rgb}{0,0.6,0}
\definecolor{codegray}{rgb}{0.5,0.5,0.5}
\definecolor{codepurple}{rgb}{0.58,0,0.82}
\definecolor{backcolour}{rgb}{1.0,1.0,1.0}

\lstdefinestyle{mystyle}{
    commentstyle=\color{codegreen},
    keywordstyle=\color{magenta},
    numberstyle=\tiny\color{codegray},
    stringstyle=\color{codepurple},
    basicstyle=\ttfamily\footnotesize,
    breakatwhitespace=false,         
    breaklines=true,                 
    captionpos=b,                    
    keepspaces=true,                 
    numbersep=0pt,                  
    showspaces=false,                
    showstringspaces=false,
    showtabs=false,                  
    tabsize=2,
    linewidth=.99\textwidth,
    xleftmargin=0.01cm
}
\usepackage{mathtools}

\newcommand{\cmark}{\text{\ding{51}}}
\newcommand{\xmark}{\text{\ding{55}}}

\newcommand{\ProbOpr}[1]{\mathbb{#1}}

\newcommand{\expect}[2]{%
\ifthenelse{\equal{#2}{}}{\ProbOpr{E}_{#1}}
{\ifthenelse{\equal{#1}{}}{\ProbOpr{E}\left[#2\right]}{\ProbOpr{E}_{#1}\left[#2\right]}}} 
\newcommand{\var}[2]{%
\ifthenelse{\equal{#2}{}}{\ProbOpr{VAR}_{#1}}
{\ifthenelse{\equal{#1}{}}{\ProbOpr{VAR}\left[#2\right]}{\ProbOpr{VAR}_{#1}\left[#2\right]}}} 

\makeatletter
\DeclareRobustCommand\onedot{\futurelet\@let@token\@onedot}
\def\@onedot{\ifx\@let@token.\else.\null\fi\xspace}

\def\eg{\emph{e.g}\onedot} 
\def\ie{\emph{i.e}\onedot}

\def\etal{\emph{et al}\onedot}
\makeatother

\newcommand{\eat}[1]{{}}

\ifCLASSOPTIONcompsoc
  \usepackage[caption=false,font=normalsize,labelfont=sf,textfont=sf]{subfig}
\else
  \usepackage[caption=false,font=footnotesize]{subfig}
\fi

\makeatletter
\def\ps@IEEEtitlepagestyle{
  \def\@oddfoot{\mycopyrightnotice}
  \def\@evenfoot{}
}
\def\mycopyrightnotice{
  {\footnotesize
  \begin{minipage}{\textwidth}
  \centering
  Copyright~\copyright~2023 IEEE. Personal use of this material is permitted. However, permission to use this \\ material for any other purposes must be obtained from the IEEE by sending a request to pubs-permissions@ieee.org.
  \end{minipage}
  }
}

\begin{document}
\title{Boosting Few-shot Fine-grained Recognition with Background Suppression and Foreground Alignment}

\author{Zican~Zha, Hao~Tang, Yunlian~Sun, and Jinhui~Tang,~\IEEEmembership{Senior~Member,~IEEE}

 \thanks{Z. Zha, H. Tang, Y. Sun, and J. Tang are with the School of Computer Science and Engineering, Nanjing University of Science and Technology, Nanjing 210094, China. E-mail: \{zhazican, tanghao0918, yunlian.sun, jinhuitang\}@njust.edu.cn. \emph{Corresponding Author: Yunlian Sun}. DOI: \url{https://doi.org/10.1109/TCSVT.2023.3236636}
 }}

\markboth{IEEE Transactions on Circuits and Systems for Video Technology, 2023}%
{Shell \MakeLowercase{\textit{et al.}}: Bare Demo of IEEEtran.cls for IEEE Journals}

\maketitle

\begin{abstract}
Few-shot fine-grained recognition (FS-FGR) aims to recognize novel fine-grained categories with the help of limited available samples. Undoubtedly, this task inherits the main challenges from both few-shot learning and fine-grained recognition. First, the lack of labeled samples makes the learned model easy to overfit. Second, it also suffers from high intra-class variance and low inter-class differences in the datasets. To address this challenging task, we propose a two-stage background suppression and foreground alignment framework, which is composed of a background activation suppression (BAS) module, a foreground object alignment (FOA) module, and a local-to-local (L2L) similarity metric. Specifically, the BAS is introduced to generate a foreground mask for localization to weaken background disturbance and enhance dominative foreground objects. The FOA then reconstructs the feature map of each support sample according to its correction to the query ones, which addresses the problem of misalignment between support-query image pairs. To enable the proposed method to have the ability to capture subtle differences in confused samples, we present a novel L2L similarity metric to further measure the local similarity between a pair of aligned spatial features in the embedding space. What's more, considering that background interference brings poor robustness, we infer the pairwise similarity of feature maps using both the raw image and the refined image. Extensive experiments conducted on multiple popular fine-grained benchmarks demonstrate that our method outperforms the existing state of the art by a large margin. The source codes are available at:~\url{https://github.com/CSer-Tang-hao/BSFA-FSFG}. 
\end{abstract}


\IEEEpeerreviewmaketitle

\section{Introduction}

\IEEEPARstart{F}{ew}-shot learning (FSL)~\cite{snell2017prototypical,hou2019cross,tang2020blockmix,FuZWFJ20,FuFJ21} has received widespread attention in the computer vision and multimedia fields because it mimics the ability of humans to learn new concepts with few available examples. Compared with the traditional classification paradigm, FSL does not depend on the large scale of labeled datasets and can be easily applied to many real-world scenarios where only very sparse training samples are available. Fine-grained recognition (FGR)~\cite{lin2015bilinear,wei2018mask,zheng2019looking,zhang2021multi,wu2021object}, a popular and challenging problem, aims to recognize images of multiple sub-categories belonging to a super-category (\eg, birds, dogs, cars). Considering the manual annotation for fine-grained images requires domain-specific knowledge, it is labor- and time-consuming to collect high quality and fully labeled large-scale datasets~\cite{YanTSL018,WangLFL22,Wang0L022}, thus FGR is a suitable application scenario of FSL. In this paper, we will study a more challenging and practical task, namely few-shot fine-grained recognition (FS-FGR), which aims to recognize fine-grained objects under few-shot settings. 

\begin{figure}[!t]
		\centering
        \includegraphics[scale=0.34]{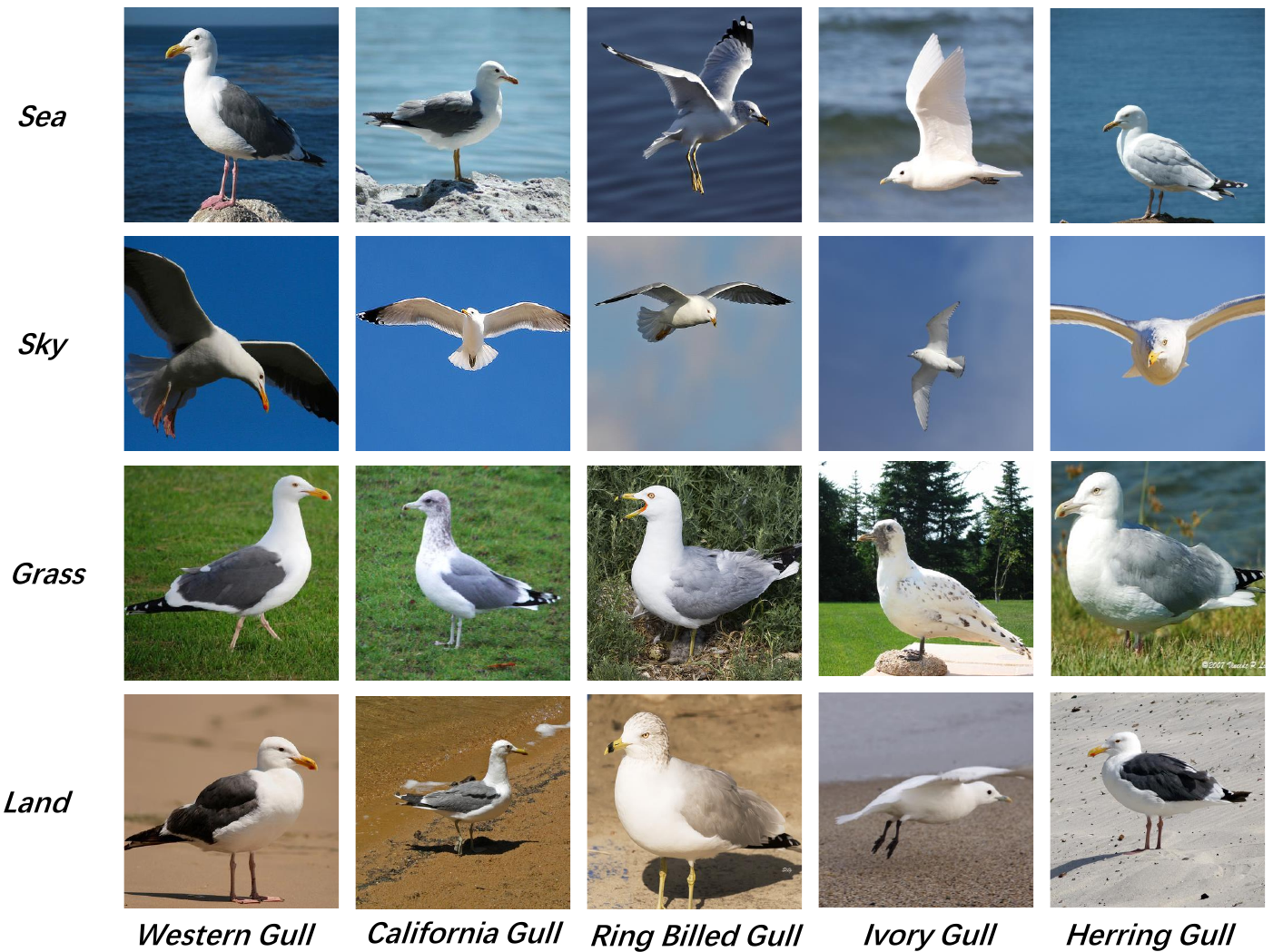}
        \vspace{-2mm}
		\caption{Key challenges of FS-FGR task. 
        Here, rows represent different backgrounds and columns represent different species. Observing horizontally, different sub-categories have low \textit{inter-class variance}. Observing longitudinally, every sub-category has high \textit{intra-class variance}. 
		}
		\label{fig:similar}
\end{figure}

\begin{figure}[tb!]
    \centering
    \includegraphics[scale=0.5]{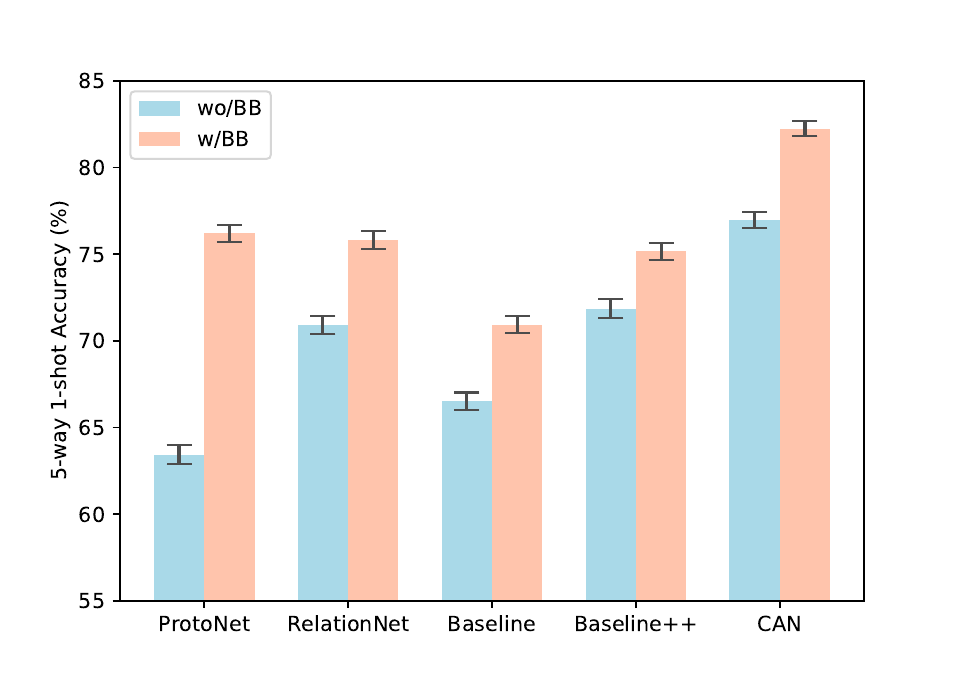}
    \vspace{-2mm}
    \caption{Performance comparison of five models with and without human-annotated bounding boxes on CUB-$200$-$2011$~\cite{wah2011caltech}: ProtoNet~\cite{snell2017prototypical}, RelationNet~\cite{sung2018learning}, Baseline~\cite{chen2019closerfewshot}, Baseline++~\cite{chen2019closerfewshot}, CAN~\cite{hou2019cross}. ``w/ BB'' denotes using the cropped images as input, where the raw images are pre-processed by human-annotated bounding boxes in advance.
    }
    \label{fig:w/oBBl}
    \vspace{-4mm}
\end{figure}

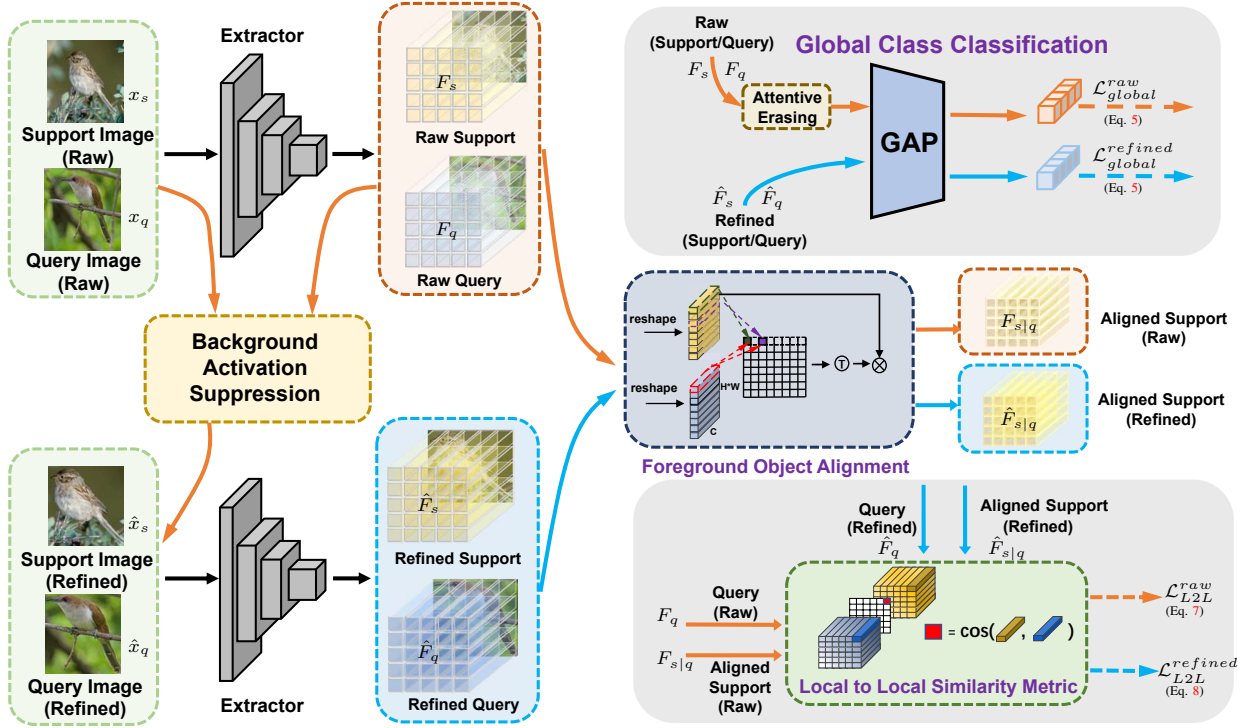
\begin{figure*}[tb!]
    \centering
    \begin{overpic}[width=0.9 \linewidth]{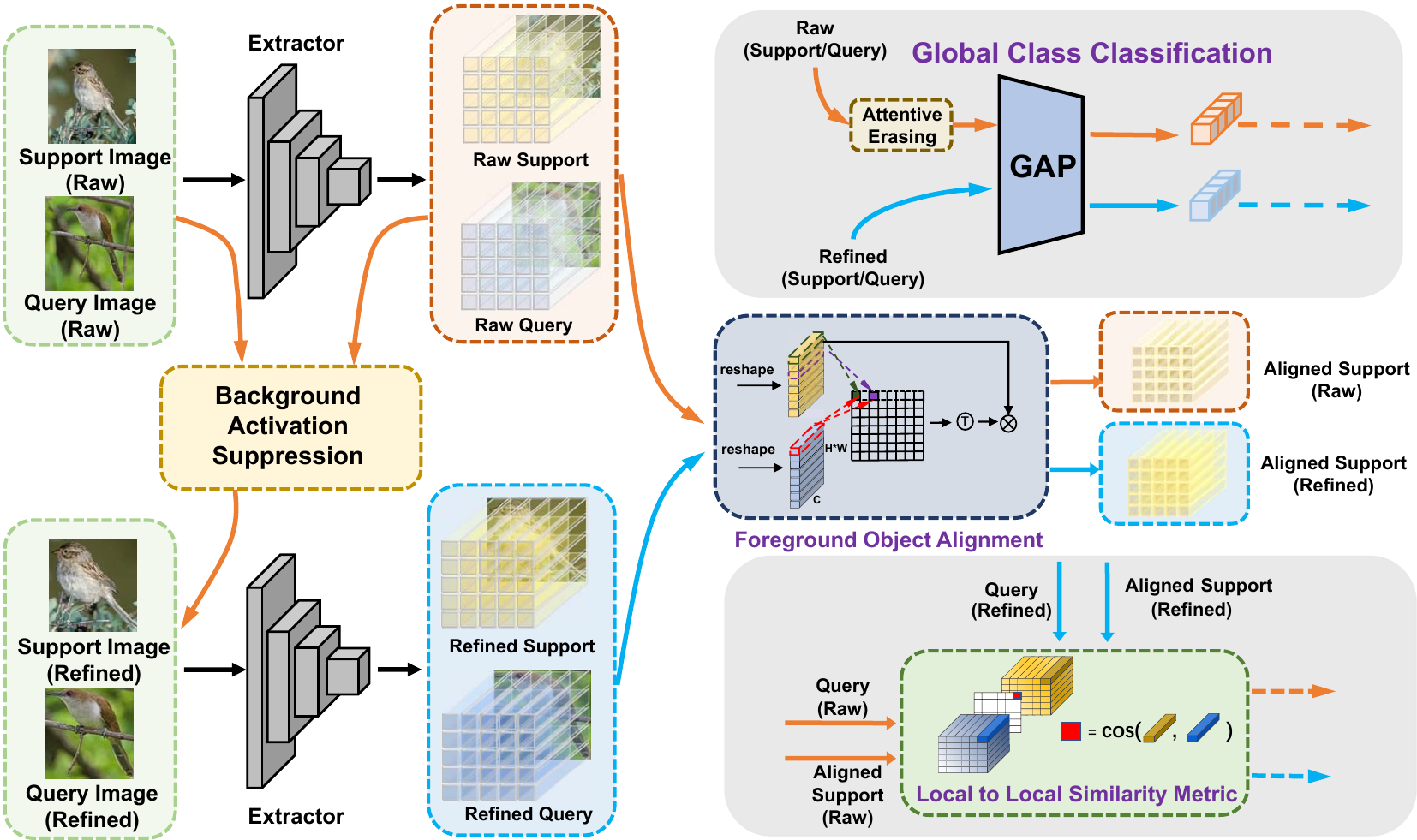}
    \put(10, 51.){\scriptsize {${x}_{s}$}}
    \put(10, 41.){\scriptsize {${x}_{q}$}}
    \put(10, 16.){\scriptsize {$\hat{x}_{s}$}}
    \put(10, 6.){\scriptsize {$\hat{x}_{q}$}}
    \put(35, 52.){\scriptsize {${F}_{s}$}}
    \put(35, 40.){\scriptsize {${F}_{q}$}}
    
    \put(55.5, 53.){\scriptsize {${F}_{s}$}}
    \put(58.5, 53.){\scriptsize {${F}_{q}$}}
    \put(57.5, 42.5){\scriptsize {$\hat{F}_{s}$}}
    \put(61.5, 42.5){\scriptsize {$\hat{F}_{q}$}}
    
    \put(33.5, 17.5){\scriptsize {$\hat{F}_{s}$}}
    \put(33.5, 5.5){\scriptsize {$\hat{F}_{q}$}}
    \put(81, 32.5){\scriptsize {${F}_{s|q}$}}
    \put(81, 24.5){\scriptsize {$\hat{F}_{s|q}$}}
    
    \put(53, 5.){\scriptsize {${F}_{s|q}$}}
    \put(53, 8.5){\scriptsize {${F}_{q}$}}
    \put(71, 14){\scriptsize {$\hat{F}_{q}$}}
    \put(80, 14){\scriptsize {$\hat{F}_{s|q}$}}
    
    \put(88.5, 51.5){\scriptsize {$\mathcal{L}^{raw}_{global}$}}
    \put(89.5, 49){\tiny{(Eq.~\ref{loss global1})}}
    \put(88.5, 46){\scriptsize {$\mathcal{L}^{refined}_{global}$}}
    \put(89.5, 43.5){\tiny{(Eq.~\ref{loss global1})}}
    \put(94.4, 10.5){\scriptsize {$\mathcal{L}^{raw}_{L2L}$}}
    \put(94.5, 9){\tiny{(Eq.~\ref{loss_local_1})}}
    \put(93.8, 4){\scriptsize {$\mathcal{L}^{refined}_{L2L}$}}
    \put(94.5, 2.5){\tiny{(Eq.~\ref{loss_local_2})}}
    \end{overpic}
    \caption{The framework of our proposed method. Note that the red line is the process of the raw stage, and the blue line is the process of the refined stage. Note that the input of the refined stage is generated by the background activation suppression module based on the input of the raw stage. Both stages have the same structure and all parameters are shared except for the global classifiers. Each stage consists of a foreground object alignment (FOA) module, a global classifier, and a similarity comparator based on L2L. 
    In the inference process, we calculate the similarity between the query-support aligned image pair in both the raw stage and the refined stage, and the similarities of both image pairs are integrated to support the final decision.
    }
    \label{fig:network}
    \vspace{-2mm}
\end{figure*}

Recently, to effectively learn from limited data, researchers have explored many general FSL algorithms for generic categories.
In general, the main FSL methods can be roughly divided into two groups, \ie, optimization-based methods~\cite{finn2017model,lee2019meta,cai2018memory,zhu2020multi} and metric-learning based methods~\cite{vinyals2016matching,snell2017prototypical,sung2018learning}. Optimization-based methods often adopt the ``learning to learn'' paradigm~\cite{andrychowicz2016learning} to learn a meta-learner that can generate a robust model. In this way, the model can easily generalize to a new unseen task with a few training samples. The model of metric-learning based methods normally consists of two parts: a feature embedding module and a comparator. The feature embedding module maps images to an embedding space. Then the distance (\eg, Euclidean distance and Cosine distance) between the support image and the query image is calculated directly in the embedding space. The comparator recognizes the query image based on the distance from each support image. Both methods have greatly promoted the performance of FSL on general datasets but few of them can obtain ideal performance on fine-grained datasets because of high intra-class variance and low inter-class difference. 

In order to migrate the model from general datasets to fine-grained datasets, current FS-FGR models generally attempt to capture key distinctions, but can easily ignore the negative effects of background. As shown in Fig.~\ref{fig:similar}, the habitats of birds are basically in the sky, land, sea, and grass. Although these birds are from different species, the same background makes them look very similar and difficult to classify. Therefore, removing the influence of the background (\eg, by manually annotating bounding boxes) is significant to boost current few-shot methods. In order to show the performance boost brought by manually annotated bounding boxes, we did an experiment on CUB-200-2011 benchmark~\cite{wah2011caltech}. The results are reported in Fig.~\ref{fig:w/oBBl}, which clearly show the significance of removing background influence. However, achieving better performance with human-annotated bounding boxes runs counter to the original aspiration of FSL to free people from burdensome and boring annotation tasks.  Another key point that is easily overlooked in Fig.~\ref{fig:similar} is the large variation across viewpoints of the same fine-grained object. Semantic patterns of fine-grained objects are mainly determined by visual appearance. Thus, aligning semantically relevant local regions remains non-trivial for the FS-FGR task, which could effectively reduce the negative effects of feature inconsistency caused by the changes of object pose or viewing angle. Therefore, our goal is to remove cluttered backgrounds and align semantically relevant foregrounds when only image-level labels and limited training images are available.

In this paper, to tackle these problems with only image-level labels available. We propose a two-stage background suppression and foreground alignment framework for few-shot fine-grained recognition, which can be trained in an end-to-end manner. As shown in Fig.~\ref{fig:network}, the proposed method mainly consists of four modules: a feature extractor, a Background Activation Suppression (BAS) module, a Foreground Object Alignment (FOA) module, and a Local to Local (L2L) similarity metric. Firstly, a feature extractor is used to extract the features for subsequent matching and localization learning. The BAS aims to generate a foreground mask map for localization based on the activation maps since the position in the activation map with a higher value are often where the interesting parts are located. Specifically, without adding extra trainable parameters, we only process the feature map to generate object location coordinates, and the BAS is supervised by the global classification loss. With the help of the generated bounding box information, we further obtained the finer scale of the object image by cropping and zooming in to remove the cluttered background. Different from the conventional method, we incorporate both the original images and the refined ones obtained by BAS into the model for learning, which can effectively generate fine-grained tailored representations for few-shot recognition.
Secondly, following DN4~\cite{li2019revisiting}, we use local descriptors to avoid losing spatial information and subtle features. Thus, to alleviate the misalignment between the support image and the query image in the embedding space due to variations in the scale, position, and posture of the foreground objects in the image, an object-aware FOA is introduced to transform support features with respect to query features by calculating a semantic correlation matrix. Thirdly, we propose an L2L similarity metric to measure the local similarity between the aligned spatial features of a given pair of samples. Experiments show that the best results are obtained by combining the similarity of raw and refined support-query pairs. Extensive experiments on three common benchmarks (\ie, CUB-$200$-$2011$~\cite{wah2011caltech}, StanfordDogs~\cite{khosla2011novel}, and StanfordCars~\cite{krause20133d}) confirm the effectiveness of the proposed method.
Our main contributions can be summarized as follows:
\begin{itemize}
\item We empirically reveal two crucial points to significantly improve the performance of the FS-FGR task,  \ie, weakening background disturbance and aligning foreground response.
\item We develop a novel two-stage weakly-supervised framework to address the FS-FGR task in a meta-learning fashion. The proposed paradigm can be optimized in an end-to-end manner with only the image-level label available.
\item Three well-designed modules are instantiated in our framework to achieve the background suppression and foreground alignment, \ie, background activation suppression module, foreground object alignment module, and local-to-local similarity metric.
\item We conduct comprehensive experiments on three popular fine-grained datasets, and our proposed model achieves state-of-the-art performance. On basis of the empirical evaluation of these datasets, we also provide insights for the elegant implementation of our solution.
\end{itemize}

The remainder of this paper is organized as follows. Sect.~\ref{set2} introduces the related work, including Few-shot Learning, Fine-grained Recognition, and Few-shot Fine-grained Recognition. Sect.~\ref{set3} gives the definition of the problem, the proposed framework, and the details of each module. Sect.~\ref{set4} presents the introduction of the datasets, the experimental results and analysis in detail, and some visualizations. The paper is concluded in Sect.~\ref{Conclude}.

\section{Related Work}\label{set2} 

\subsection{Few-shot Learning}
Few-shot learning (FSL), which aims to learn a well-generalized model with only a few samples, has attracted intensive attention in recent years. As one of the standard benchmarks in meta-learning, existing work on FSL can be roughly categorized into three groups according to their innovation: metric-based methods, optimization-based methods, and augmentation-based methods.
Metric-based methods~\cite{vinyals2016matching,snell2017prototypical,sung2018learning} aim to learn a suitable model that projects all images to a metric space, and then predicts query images according to their distances to the labeled support samples. Optimization-based methods~\cite{finn2017model,rusu2018meta,lee2019meta} follow the ``learning to learn'' paradigm to learn a good initialization for the model or learn how to update the parameters of the model within a few iterations for quick adaption. Augmentation-based methods~\cite{ZhangCGBS18,HariharanG17,AFHN} aim to learn a generator from base categories and use it with additional techniques to increase the number of novel samples or features for data augmentation. 
In this work, we are particularly interested in the metric-based methods owing to their intuitive and effective characteristics, which are favored by state-of-the-art methods on many tasks.

Metric-based methods mainly focus on two key factors, \ie,~\textit{feature extraction}~\cite{li2019revisiting,tang2020blockmix} and \textit{similarity measurement}~\cite{hou2019cross,zhang2020deepemd,HaoHCT22}. The former is responsible for mapping the samples to an ideal embedding space, where the samples belonging to the same category are close together and the samples belonging to different categories are far apart. The latter serves as a criterion for the nearest neighbor search to perform recognition. 
However, most of the existing metric-based FSL methods~\cite{tang2020blockmix,kang2021relational,0002HGD21,ZhangLC20} mainly focus on learning a discriminative global embedding for coarse-grained generic object recognition, which are less suitable to address the few-shot fine-grained recognition (FS-FGR) task well. The main reason is that fine-grained recognition requires to emphasis on distinguished appearance details of local parts, but global features are incapable of mining spatially local discriminative parts, especially under the few-shot scenario. By comparison, the proposed method proposes a foreground alignment module to emphasize the semantically correlated parts of the input support-query pair via spatially aligning dense local features for FS-FGR.

\subsection{Fine-grained Recognition}
Fine-grained recognition~\cite{Wei2021survey} aims to recognize multiple subordinate categories belonging to the same super-category~(\eg,~bird species~\cite{wah2011caltech}, car models~\cite{khosla2011novel} and dog breeds~\cite{krause20133d}), which has gained much attraction of the research community. Compared to the coarse-grained recognition~\cite{LiYYT20,LiDYTQ22,YanZXZY22}, fine-grained recognition is a more challenging task, since fine-grained objects are usually distinguished by local feature variations or subtle feature differences.
Earlier work~\cite{ZhangDGD14,lam2017fine} mainly relies on part annotations or hand-crafted bounding boxes to locate discriminative specific parts, which exacerbates the cost of prior information or additional annotations. 
Benefiting from the remarkable progress of powerful deep neural networks and large-scale annotated datasets, some deep learning-based approaches~\cite{zheng2019looking,zhang2021multi, DingMWXCSWL21} attempt to learn discriminative features or locate the distinguishable parts in a weakly supervised manner, where only image-level category labels are available. These methods can be roughly categorized into two main groups: \textit{feature encoding}-based methods~\cite{lin2015bilinear,GaoBZD16,HuYZCZ21} and \textit{part localization}-based methods~\cite{FuZM17,WeiXWS18,HePZ19}. 
For example, Guo~\etal~\cite{TBALNet2021} introduce  a lightweight attention module to localize critical regions and learn fine-grained feature representation.
More recently, MMAL~\cite{zhang2021multi} and AP-CNN~\cite{DingMWXCSWL21} propose to first find strong discriminative regions, and then re-input the original image or feature maps to the network by cropping and resizing them, which further enhance the object representation’s discrimination. The above novel paradigm also serves as some inspiration for our proposed method.

Despite the success brought by the above methods, they still rely on large-scale datasets, which are less practical in real scenarios, due to the large-scale annotated datasets being hard to attain in some cases.
Compared with previous work, we study fine-grained recognition under the challenging few-shot learning setting, namely few-shot fine-grained recognition (FS-FGR), where only a few labeled samples are provided to recognize novel fine-grained categories.

\subsection{Few-shot Fine-grained Recognition}
Recently, inspired by the rapid development of FSL, there are some emerging work~\cite{wei2019piecewise, li2020bsnet, TANG2022108792} that starts to explore the intractable few-shot fine-grained recognition (FS-FGR) task, where the goal is to distinguish the novel sub-categories in a super-category given a limited number of labeled samples.
Wei~\etal~\cite{wei2019piecewise} propose the first solution for FS-FGR by introducing a bilinear pooling network to encode features and multiple sub-classifiers to predict the label.
LRPABN~\cite{HuangZZXW21} proposes to learn multiple transformations to match the embedded input pairs.
MattML~\cite{zhu2020multi} introduces multi-attention mechanisms to learn a task-specified classifier initialization, which is helpful to capture diverse discriminative parts. 
%
%
%
However, the above methods tackle the FS-FGR task by learning and matching diverse and informative global features, which follow the same perspective as the generic FSL methods. Limited efforts have been devoted to distinguishing the subtle difference from the perspective of dense local features.
In contrast, CPSN~\cite{tian2021coupled} introduces two coupled branches to compute the similarity scores between input pairs from patch level to capture subtle and local differences.

More recently, considering object location differences in the support and query set cause performance degradation, some works have introduced feature alignment to solve the FS-FGR problem, which has achieved very inspiring results. 
TOAN~\cite{huang2021toan} proposes a target-oriented matching mechanism to learn explicit feature transformations to reduce the intra-class variance. 
Wu~\etal~\cite{WuLGJ19} introduce a dual correlation attention mechanism to explore position-wise relationships of two compared features to capture their semantic information.
Wu~\etal~\cite{wu2021object} propose an object-aware long-short-range spatial alignment method to align features between various support and query features. 
Although the above cross-correction attention-based workflow seems to make sense, they are all pretty sophisticated and obscure.

We can see that existing FS-FGR methods usually focus on how to extract discriminative features and how to strengthen the relation matching process of input pairs by different metric functions. However, inherent characteristics (\eg,~pose variations and similar backgrounds) of fine-grained images are also important factors for the FS-FGR task that are often ignored by existing FS-FGR methods.
In this paper, motivated by aforementioned work in the FS-FGR community, we argue that \textbf{background suppression} and \textbf{foreground alignment} are essential to recognize fine-grained objects, and they have been demonstrated to be very appropriate for the FS-FGR task and could significantly improve the performance.

We are aware that recently AGPF~\cite{TANG2022108792} proposed an attention-guided refinement strategy to enhance the dominative object and conducted a two-stage meta-learning framework to capture attention-guided pyramidal features. Although AGPF also adopts a two-stage framework to weaken the background disturbance, our approach obviously differs from AGPF in these aspects, at least: (1) Our method encodes each input into a set of dense local features and introduces a feature alignment technique for fine-grained objects, which is more suitable for capturing subtle appearance differences of local parts than a single informative global feature used in AGPF. (2) The attention-guided refinement proposed in AGPF utilizes sophisticated attention mechanisms to locate the dominative foreground. However, ours only relies on the feature activation map to predict and disentangle the foreground object and background noise, which is parameter-free. (3) AGPF aggregates complementary features discovered from different scales and stages to form a unified global embedding, but our method computes the spatial similarity between aligned feature pairs in two stages to jointly support the final decision. (4) Compared with AGPF, our results are significantly better and have significant advantages in efficiency.

\section{Methodology}
\label{set3}
\subsection{Problem Formulation}
In the standard setting of few-shot learning, given a large-scale labeled dataset with base categories $\mathcal{C}_{base}$, the goal is to learn a standard embedding network that can be easily adapted to a family of unseen tasks $\{\mathcal{T}\}_{1}^{n}$ with novel categories $\mathcal{C}_{novel}$, where $\mathcal{C}_{base} \cap \mathcal{C}_{novel} = \emptyset$.
Considering a specific image recognition task $\mathcal{T}$, it usually contains a \textit{support set} $\mathbf{S}$ and a \textit{query set} $\mathbf{Q}$ where both are sampled from the same novel categories $\mathcal{C}_{novel}$, our goal is to determine the category of samples in the unlabeled $\mathbf{Q}$ according to the labeled $\mathbf{S}$. In this work, we study a ``$N$-way $K$-shot'' task, where the support set $\mathbf{S}$ contains $N$ categories and each category has $K$ images. Concretely, $N$ is set to $5$ and $K$ is set to $1$ or $5$ in most previous studies, and so do we. To have a good generalization on novel categories $\mathcal{C}_{novel}$ and avoid the overfitting on base categories $\mathcal{C}_{base}$, we also adopt the episodic training mechanism~\cite{vinyals2016matching} widely used in most literature.

\subsection{The Proposed Framework}
\label{framework}
Given a ``$N$-way $K$-shot'' fine-grained recognition task, we need to divide it into a support set $\mathbf{S}=\{ (x_{i}^{j}, y_{i}^{j})~|~i= 1\cdots N, j=1\cdots K\}$ and a query set $\mathbf{Q}=\{ (x_{i}, y_{i})~|~i= 1\cdots |\mathbf{Q}|\}$, and then map all samples in a task into an embedding space by a feature extractor $\Theta$, \ie, $F_{i}=\Theta(x_{i})\in\mathbb{R}^{c\times h\times w}$, where $x_{i}\in \mathbf{S}\cup \mathbf{Q}$. 
Here, $c$ is the number of channels, $h$ and $w$ denote the size of the feature map. 
Note that the proposed method borrows the idea of metric learning and needs to match each query sample to support prototypes by the nearest neighbor strategy. Therefore, the $i_{{th}}$ support prototype $\mathbf{C}_{i}$ can be formulated as $\mathbf{C}_{i}=\frac{1}{K}\sum_{j=1}^{K}\Theta(x_{i}^{j})$, where $x_{i}^{j}$ is a sample labeled with category $i \in \{1,\cdots,N\}$.

The overview of the proposed method is shown in Fig.~\ref{fig:network}. Given one support sample $x_{s}\in \mathbf{S}$ and one query sample $x_{q}\in \mathbf{Q}$, we first extract the corresponding feature maps $F_{s}$ and $F_{q}$, respectively.
Next, the \textit{background activation suppression} module is used to remove the cluttered background based on the activation value of the feature map and then zoom in the cropped region to the size of the raw image. $\hat{x}_{s}$ and $\hat{x}_{q}$ represent the refined support image and the refined query image, respectively. We extract the refined feature maps $\hat{F}_{s}$ and $\hat{F}_{q}$  from the same backbone \(\Theta\). Here, the feature extractors in the upper and lower branches are parameter-shared. Subsequent steps first use the proposed \textit{foreground object alignment} module to align two semantic feature pairs (\ie, $\{F_{s}, F_{q}\}$ and $\{\hat{F}_{s}, \hat{F}_{q}\}$) spatially, and then compute the total spatial similarity between two aligned pairs by introducing the proposed ``\textit{local to local}'' similarity metric. Finally, we integrate the similarity scores of both pairs to classify the novel query set $\mathbf{Q}$.

\begin{figure}[t!]
\centering
\includegraphics[width=0.9 \linewidth]{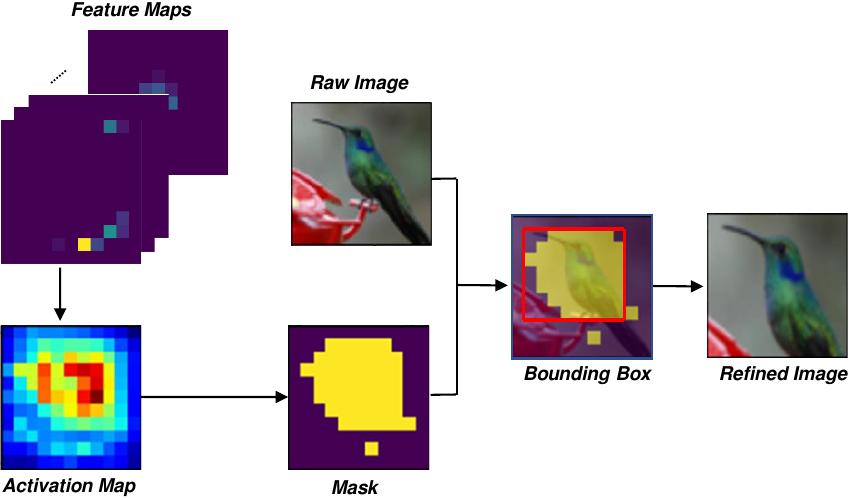}
\vspace{-2mm}
\caption{Background Activation Suppression (BAS) Module. For the input feature, BAS first generates a class-agnostic activation map for localization. The foreground prediction map is then obtained by suppressing the background activation value.}
\label{fig:BAS}
\vspace{-4mm}
\end{figure}
\subsection{Background Activation Suppression Module}
\label{BAS}
\textbf{Motivation.}
As we all know, traditional fine-grained recognition is a challenging problem due to the small inter-category variations and the large intra-category variations that exist in the fine-grained dataset. As shown in Fig.~\ref{fig:similar}, the visual patterns of different sub-categories are similar to some extent, and the backgrounds in the same sub-categories may be different but appear slightly similar among different sub-categories. Therefore, the image background plays a negative role in the fine-grained recognition task, especially in the few-shot scenario. Inspired by the weakly-supervised localization methods~\cite{wei2018mask,zhang2021multi} for the fine-grained task, the proposed method attempts to eliminate the negative effect of background by the BAS module, without additional annotations except for image-level annotation. BAS directly generates class-agnostic activation maps to disentangle the foreground object and background in an image.

For example, given an input query image $x_{q}$, the corresponding feature map $F_{q}=\Theta(x_{q})\in\mathbb{R}^{c\times h\times w}$ is generated by the feature extractor. Note that each channel of a feature map can be regarded as a pattern detector~\cite{zeiler2014visualizing}, denoted by $f_{i}$ ($i=1,\cdots, c$). 
If a certain spatial position in the feature map has high activation for most channels, it is likely to correspond to an informative region. Therefore, we obtain the activation map $A_{F}\in\mathbb{R}^{h\times w}$ by aggregating $F$ along the channel dimension as $A_{F} = \sum_{i=1}^{c} f_{i}$. Then, we introduce an adaptive threshold $\theta_{A_{F}}$ for the activation map $A_{F}$ to determine which position is a part of the object regions, \ie, $\theta_{A_{F}}=\frac{\sum_{i=1}^{w} \sum_{j=1}^{h} A_{F}(i, j)}{h\times w}$. Finally, we generate a foreground mask $M_{A_{F}} \in \{0,1\}^{h\times w}$ by comparing each element of $A_{F}$ with threshold $\theta_{A_{F}}$. Concretely, for a particular position $(i,j)$, if the activation value of $A_{F}(i, j)$ is larger than the threshold $\theta_{A_{F}}$, the corresponding $M_{A_{F}}(i,j)$ is  set to $1$ otherwise $0$. Formulaically,
\begin{equation}
    \begin{aligned}
        \label{mask_map}
        M_{A_{F}}(i,j)&=
		\begin{cases}
			1,& {if}~A_{F}(i, j)>\theta_{A_{F}} \\
			0,& {otherwise.}
		\end{cases}
    \end{aligned}
\end{equation}
Here, we search the largest connected component of the generated $M_{A_{F}}$ to obtain the smallest bounding box for the foreground prediction. Based on the location coordinates, we can perform background suppression by cropping interesting foreground on raw image $x_{{q}}$ and enlarging it to the same size as $x_{q}$ for subsequent classification, denoted as $\hat{x}_{q}$. The processing steps of the BAS module are illustrated in Fig.~\ref{fig:BAS}.

\begin{figure}[t!]
\centering
\includegraphics[scale=0.32]{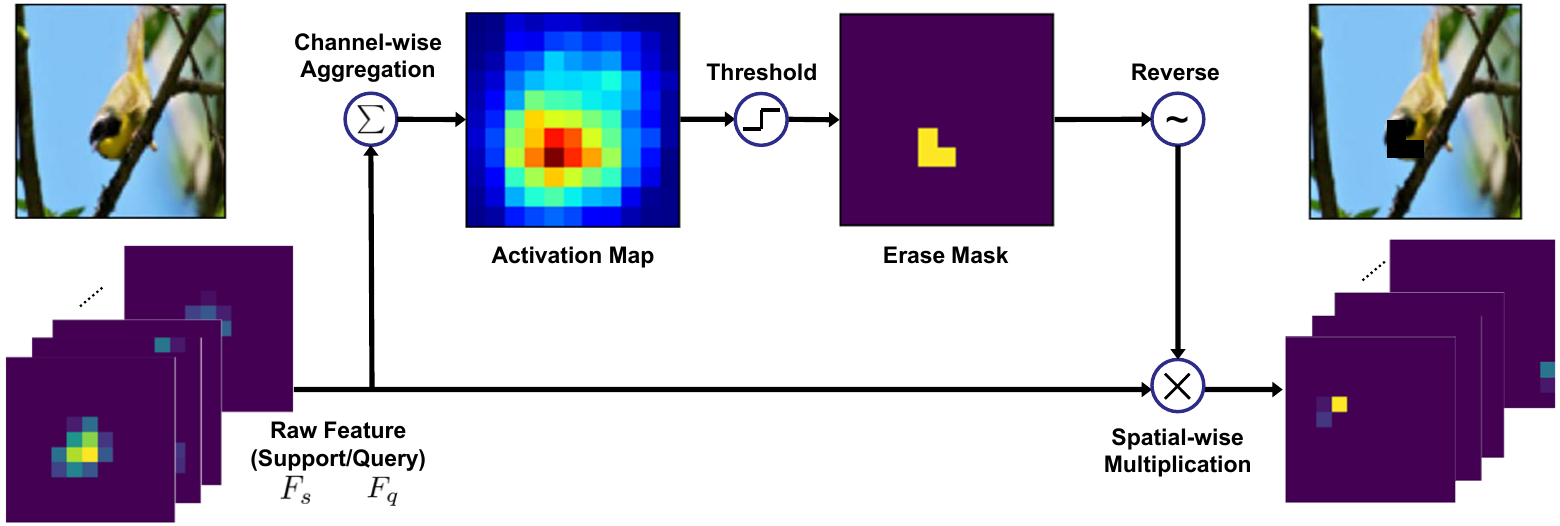} 
\vspace{-2mm}
\caption{Attentive erasing diagram. In the raw stage, to encourage the network to minimize the ratio of background activation and overall activation generated by the entire image, we erase the region of raw feature with activation higher than a threshold before being fed into the global classifier, which is helpful to explore the entire extent of the foreground.}
\label{fig:erasing}
\vspace{-4mm}
\end{figure}

Due to the unbalanced distribution between the foreground features and background features, fine-grained recognition models incline to identify patterns from small discriminative object regions. 
To address this issue, attentive erasing strategy~\cite{HouJWC18,WeiFLCZY17} as an efficient solution has been widely used in weakly supervised object localization. However, these methods have limited potential to derive complete foreground activation maps of unseen objects.
To further improve the completeness of the foreground prediction map with image-level labels, we propose to promote the learning of BAS by erasing the prominent regions of the feature map with activation higher than a threshold as shown in Fig.~\ref{fig:erasing}. Firstly, we generate an activation map $M_{att}\in\mathbb{R}^{h\times w}$ as same as $A_{F}$ in the BAS module by 
aggregating raw feature (\ie~$F_{s}$ or $F_{q}$) along the channel dimension. Then, the spatial distribution could be searched for the most discriminative part based on $M_{att}$, due to the intensity of each position being proportional to the discriminative power. To obtain the erase mask $M_{era}\in\mathbb{R}^{h\times w}$, a threshold $\theta_{M_{F}}$ is introduced by setting a ratio $\gamma$ of maximum intensity of $M_{att}$. Thus, $M_{era}$ is produced by setting each pixel to $1$ if it is larger than $\theta_{M_{F}}$ and $0$ if it is smaller, and the size of the erased region decreases as $\gamma$ increases. For the erasing threshold, we set hyper-parameter $\gamma$ to $85\%$ in our recommended settings. Finally, the mask $M_{era}$ is applied to the input feature map by inversion and spatial-wise multiplication. This attentive erasing strategy facilitates the
generation of a meaningful class-agnostic activation map to explore the entire foreground object by indirectly increasing the foreground activation value. This also encourages the model to learn the less discriminative parts.

\begin{figure}[t!]
\centering
\includegraphics[scale=0.32]{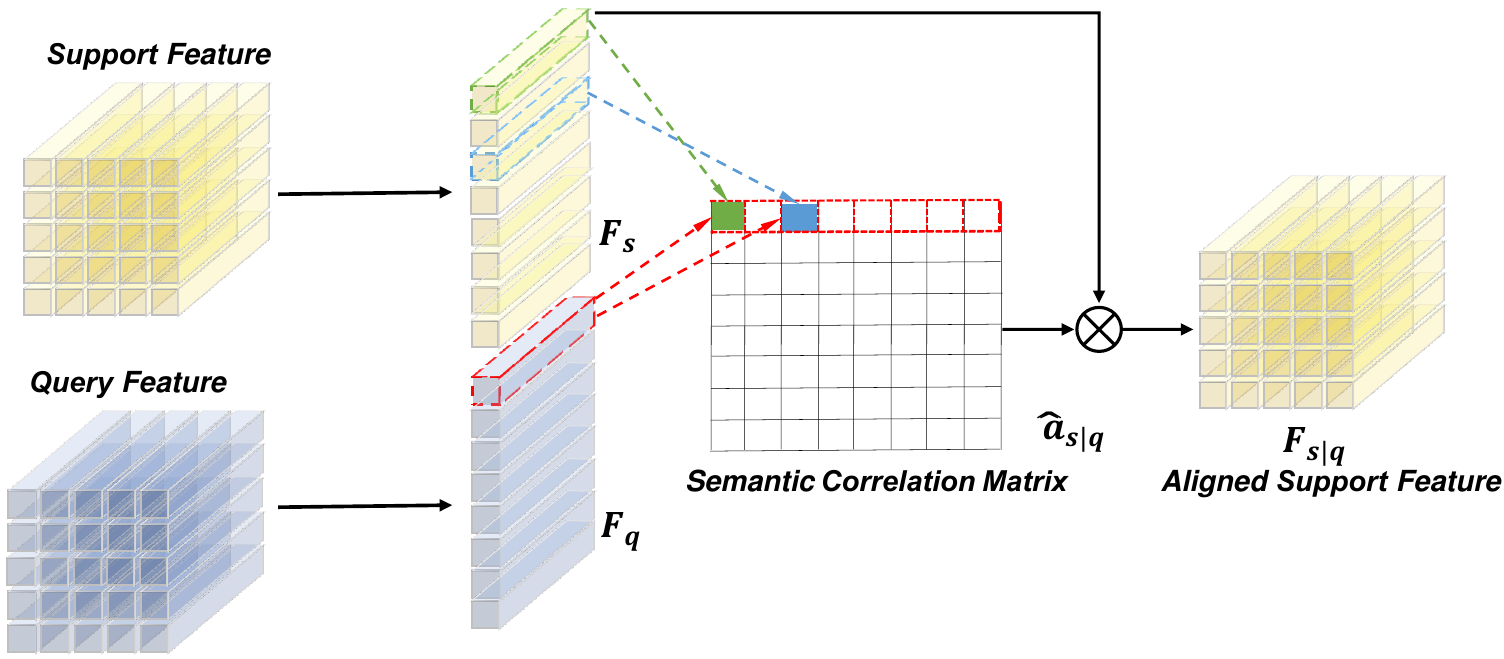} 
\vspace{-2mm}
\caption{Foreground Object Alignment (FOA) Module. The input support-query image pair is initially misaligned, due to the arbitrary pose and position variations of foreground between the two fine-grained images.
By leveraging the proposed FOA module, semantic features of the original support image could be aligned \textit{w.r.t} the query image.}
\label{fig:FOA}
\vspace{-4mm}
\end{figure}

\subsection{Foreground Object Alignment Module}
\label{Feature Alignment Module}
\textbf{Motivation.}
Although global embedding generated by global average pooling or global max pooling operations is capable of capturing discriminative information in general few-shot learning tasks, it might suppress some local discriminative characteristics while overemphasizing the irrelevant background features for the fine-grained object~\cite{YSL2022}, hurting the FG-FSR performance. 
As illustrated in Fig.~\ref{fig:similar}, we can see that capturing the subtle differences in discriminative parts remains non-trivial to distinguish similar sub-categories. To this end, our proposed method attempts to exploit dense local features to calculate the similarity between a given pair of samples, which can further capture spatial structure information of the foreground for FG-FSR.
In addition, the lower intra-category visual consistency of fine-grained objects usually causes different similarities in both global appearance and various local regions. Therefore, an ideal solution to FS-FGR should not only be sensitive to the subtle discrepancy among instances from different sub-categories but also be invariant to the arbitrary poses, scales, and appearances of the same subcategory. 
In light of this, our proposed method goes two steps further to compute the similarity of each support-query feature pair: (1) aligning semantic features between a query image and various support images according to their correlations; (2) modeling the similarity measurement between each aligned support-query pair as a local-to-local (patch-to-patch) matching process.

Specifically, before locally comparing a support feature $F_{s}\in \mathbb{R}^{c\times h\times w}$ and query feature $F_{q}\in \mathbb{R}^{c\times h\times w}$, we first need to address the issue of semantic misalignment, due to intra-class variance (\eg, object posture and position variation) and cluttered background. In detail, we first reshape a pair of spatial features from $\mathbb{R}^{c\times h\times w}$ to $\mathbb{R}^{hw\times c}$, \ie, $F_{s}=\{s_i\}_{i=1}^{h\times w}$ and $F_{q}=\{q_i\}_{i=1}^{h\times w}$, where $s_{i}$ and $q_{i}\in\mathbb{R}^{c\times 1\times 1}$ represent the $i_{\text{th}}$ deep local descriptor in $F_{s}$ and $F_{q}$, respectively. Then, we use the reshaped $F_{s}$ and $F_{q}$ to compute a semantic correlation matrix $a_{s|q}\in\mathbb{R}^{hw\times hw}$, which is calculated by the cosine similarity as $a_{s|q}(i,j) = \frac{q_{i}^{\text{T}} s_{j}}{\|q_i\|\cdot\|s_j\|}$.
Furthermore, a normalized operation $\mathrm{softmax}(\cdot)$ is applied to $a_{s|q}$ row-by-row, so that the sum of each row in the normalized $a_{s|q}$ is $1$. The resulting semantic correlation matrix is formulated as follows:
\begin{equation}
\hat{a}_{s|q}=\mathrm{softmax}(a_{s|q})=\frac{\exp \left(a_{s|q} (i,j)\right)}{\sum_{m=1}^{h\times w} \exp \left(a_{s|q}(i,m)\right)}. 
\end{equation}
As shown in Fig.~\ref{fig:FOA}, we employ the FOA module to compute the semantic correlation matrix $\hat{a}_{s|q}$, which is used to align the support feature $F_{s}$ with respect to query feature $F_{q}$. Formulaically, the aligned feature $F_{s|q}$ can be computed by an alignment function $\mathcal{A}(\cdot, \cdot)$ as: 
\begin{equation}
F_{s|q} = \mathcal{A}(F_{s}, F_{q}) = \hat{a}_{s|q}\cdot F_{s}. 
\end{equation}

\begin{figure}[t!]
\centering
\includegraphics[scale=0.4]{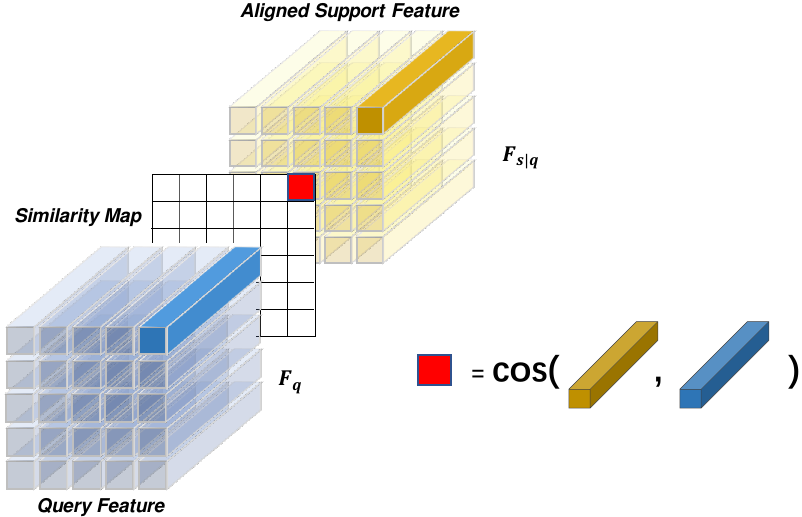} 
\vspace{-1mm}
\caption{\label{fig:L2L}Local to Local (L2L) Similarity Metric. We compute one similarity map by measuring the cosine similarity element-wisely between the two aligned feature maps of compared images.}
\vspace{-3mm}
\end{figure}

Considering that the semantic features aligned by the FOA module are no longer affected by the position deviation, we further propose a patch-level similarity metric, indicated by $L2L(\cdot, \cdot)$, which reformulates similarity measurement as a local-to-local matching process. As shown in Fig.~\ref{fig:L2L}, given a pair of aligned semantic features $\{F_{s|q}, F_{q}\}\in \mathbb{R}^{hw\times c}$, the $L2L$ metric can be formulated as a general function of $F_{s|q}$ and $F_{q}$:
	\begin{equation}
		L2L(F_{s|q}, F_{q})=\sum_{i=1}^{h}\sum_{j=1}^{w}cos(F_{s|q}^{i,j}, F_{q}^{i.j}),
	\end{equation}
where $F^{i,j}\in \mathbb{R}^{c}$ is a deep descriptor of position $(i,j)$ in the semantic feature $F$ and $cos(\cdot, \cdot)$ denotes the cosine similarity function.

\subsection{Overall Loss Function} \label{loss_function}
During the training phase, given a ``$N$-way $K$-shot'' fine-grained recognition task $\mathcal{T}$, each sample $x_{i}$ in $\mathcal{T}$ has two types of labels: $y^{g}\in\{1,2,\cdots,G\}$ and $y^{l}\in\{1,2,\cdots,N\}$, where $G$ is the overall number of base categories. Firstly, to learn a class-agnostic activation map for achieving   foreground prediction and background suppression, we use cross-entropy loss as the classification loss to explicitly explore rich inter-class relationships by learning features related to the whole base category space. As shown in Fig.~\ref{fig:network}, there are two independent global classifiers in our proposed framework to classify features of the raw image $x_{i}$ and the refined image $\hat{x}_{i}$ to its corresponding base category $y^{g}$. The corresponding global classification loss $\mathcal{L}^{raw}_{global}$ and $\mathcal{L}^{refined}_{global}$ are formulated as follows: 
\begin{equation}\label{loss global1}
\resizebox{0.9\hsize}{!}{$
		\begin{aligned} 
			\mathcal{L}^{raw}_{global}&= -\sum _{(x_{i}, y^{g})\in \mathcal{T}}y^{g}\log((\mathrm{softmax}(W_{1} \mathrm{GAP}(\Theta(x_{i}))))),\\
			\mathcal{L}^{refined}_{global}&= -\sum _{(\hat{x}_{i}, y^{g})\in \mathcal{T}}y^{g}\log((\mathrm{softmax}(W_{2} \mathrm{GAP}(\Theta(\hat{x}_{i}))))).
		\end{aligned}$}
\end{equation}
Here, $W_{1}$, $W_{2}$ are the weight parameters of the global classifiers, $\mathrm{GAP}(\cdot)$ denotes the global average pooling, $y^{g}$ is the true base category label for $x^{raw}$ and $x^{refined}$. 
Finally, the global classification loss $\mathcal{L}_{global}$ can be formulated as follow:
\begin{equation}\label{global_loss}
\mathcal{L}_{global} = \alpha \mathcal{L}^{raw}_{global} + \beta \mathcal{L}^{refined}_{global},
\end{equation}
where $\alpha$ and $\beta$ are two weighting factors.
Note that the global classification branch is only a part of the whole framework and will be activated in the training phase to constrain the generation of class-agnostic activation maps.

Our goal is to identify each query sample with a few labeled samples in the support set. We compute the similarity between the query and each support prototype and then the matching process is guided by the nearest-neighbor strategy. For task $\mathcal{T}$ with $N$ categories, given a prototype set $\{\mathbf{C}_{j}\}_{j=1}^{N}$ constructed from raw support images, the local-level few-shot loss between the feature $F_{i}$ of raw query image $x_{i} \in \mathbf{Q}$ (\ie,~$F_{i}=\Theta(x_{i})$) and the corresponding prototype $\mathbf{C}_{j}$ can be represented by 
	\begin{equation}\label{loss_local_1}
	\resizebox{0.9\hsize}{!}{$
		\mathcal{L}^{raw}_{L2L}=-\sum_{i=1}^{\lvert \mathbf{Q}\rvert} \log \frac{\exp \left(-L2L\left(\mathcal{A}(\mathbf{C}_{j}, F_{i}), F_{i}\right)\right).}{\sum_{j^{\prime}=1}^{N} \exp (-L2L\left(\mathcal{A}(\mathbf{C}_{j^{\prime}}, F_{i}), F_{i}\right))}.$}
	\end{equation}
Similar to $\mathcal{L}_{global}$ in Eq.~\ref{loss global1}, we replace the raw query image $x_{i}$ in Eq.~\ref{loss_local_1} with the refined query image $\hat{x}_{i}$ to obtain $\mathcal{L}^{refined}_{L2L}$ as follow:
\begin{equation}\label{loss_local_2}
\resizebox{0.9\hsize}{!}{$
		\mathcal{L}^{refined}_{L2L}=-\sum_{i=1}^{\lvert \mathbf{Q}\rvert} \log \frac{\exp \left(-L2L\left(\mathcal{A}(\hat{\mathbf{C}}_{j}, \hat{F}_{i}), \hat{F}_{i}\right)\right).}{\sum_{j^{\prime}=1}^{N} \exp (-L2L\left(\mathcal{A}(\hat{\mathbf{C}}_{j^{\prime}}, \hat{F}_{i}), \hat{F}_{i}\right))},$}
	\end{equation}
where $\hat{F}_{i}$ indicates the feature of refined query image $\hat{x}_{i}$ generated by BAS, and $\hat{\mathbf{C}}_{j^{\prime}}$ is the corresponding prototype computed by refined support images.
Therefore, the final local-level few-shot loss $\mathcal{L}_{local}$ can be represented as follow:
\begin{equation}\label{local_loss}
\mathcal{L}_{local} = \alpha \mathcal{L}^{raw}_{L2L} + \beta \mathcal{L}^{refined}_{L2L},
\end{equation}
where $\alpha$ and $\beta$ are two weighting factors as same as hyper-parameters in Eq.~\ref{global_loss} to control the contributions of two stages. Here, two scale factors $\alpha$ and $\beta$ are both set to 0.5 for all experiments, while optimizing them can further improve the performance.

In summary, the cooperative loss is formulated as follows:
\begin{equation}
\mathcal{L}_{total}=\mathcal{L}_{global}+\lambda \mathcal{L}_{local},
\end{equation}
where $\lambda$ is a weighting factor. In the inference phase, only the branch of local-level few-shot loss $\mathcal{L}_{local}$ is used to compute the similarity to recognize the query samples.

\begin{table*}[tb!]
	\centering
	\caption{ \label{tab:1}The accuracy (\%) of $5$-way $1$-shot and $5$-shot tasks on three popular benchmarks. We report the mean accuracy with 95\% confidence intervals on $2000$ test episodes. The best results are shown in bold. * denotes the results obtained from the original paper. We reproduce the other results under the same experimental settings according to the open source code.}
	\vspace{-2mm}
	\resizebox{0.92\linewidth}{!}{ 
		\begin{tabular}{lcccccc}
			\addlinespace
			\toprule
			\specialrule{0em}{1pt}{1pt}
			\multirow{2}{*}{ \bf Method} &\multicolumn{2}{c}{\bf CUB-200-2011} & \multicolumn{2}{c}{\bf Stanford Dogs} & \multicolumn{2}{c}{\bf Stanford Cars} \\ 
			\cline{2-7}
			\specialrule{0em}{1pt}{1pt}
			&{$5$-way $1$-shot} & {$5$-way $5$-shot}& {$5$-way $1$-shot} &{$5$-way $5$-shot}& {$5$-way $1$-shot} &{$5$-way  $5$-shot}  \\
			\midrule
			\specialrule{0em}{1pt}{1pt}
			ProtoNet~\cite{snell2017prototypical}  & {63.44} {\scriptsize $\pm$ 0.56} & {83.17} {\scriptsize $\pm$ 0.35} & {41.61} {\scriptsize $\pm$ 0.50}& {76.78} {\scriptsize $\pm$ 0.36} & {45.01} {\scriptsize $\pm$ 0.49}& {87.19} {\scriptsize $\pm$ 0.31}\\
			\specialrule{0em}{1pt}{1pt}
			RelationNet~\cite{sung2018learning}  & {70.92} {\scriptsize $\pm$ 0.54}  & {84.90} {\scriptsize $\pm$ 0.35} & {61.21} {\scriptsize $\pm$ 0.51}     & {80.27} {\scriptsize $\pm$ 0.37} & {78.04} {\scriptsize $\pm$ 0.53} & {90.03} {\scriptsize $\pm$ 0.30} \\
			\specialrule{0em}{1pt}{1pt}
			DN4~\cite{li2019revisiting}  & {64.95}  {\scriptsize $\pm$ 0.99} & {83.18} {\scriptsize $\pm$ 0.62} &  {49.70} {\scriptsize $\pm$ 0.85}  & {71.59} {\scriptsize $\pm$ 0.68}  & {75.79} {\scriptsize $\pm$ 0.84} & {94.14} {\scriptsize $\pm$ 0.35}\\
			\specialrule{0em}{1pt}{1pt}
			Baseline~\cite{chen2019closerfewshot} & {66.52}  {\scriptsize $\pm$ 0.50} & {83.31} {\scriptsize $\pm$ 0.34} &  {59.19} {\scriptsize $\pm$ 0.48}  & {78.21} {\scriptsize $\pm$ 0.34}  & {69.48} {\scriptsize $\pm$ 0.46} & {87.81} {\scriptsize $\pm$ 0.27}\\
			\specialrule{0em}{1pt}{1pt}
			Baseline++~\cite{chen2019closerfewshot} & {71.85}  {\scriptsize $\pm$ 0.55} & {82.14} {\scriptsize $\pm$ 0.62} &  {58.53} {\scriptsize $\pm$ 0.51}  & {73.12} {\scriptsize $\pm$ 0.38}  & {76.79} {\scriptsize $\pm$ 0.53} & {87.18} {\scriptsize $\pm$ 0.32}\\
			\specialrule{0em}{1pt}{1pt}
			CAN~\cite{hou2019cross}  & {76.98}  {\scriptsize $\pm$ 0.48} & {87.77} {\scriptsize $\pm$ 0.30} &  {64.73} {\scriptsize $\pm$ 0.52}  & {77.93} {\scriptsize $\pm$ 0.35}  & {86.90} {\scriptsize $\pm$ 0.42} & {93.93} {\scriptsize $\pm$ 0.22}\\
			\specialrule{0em}{1pt}{1pt}
			TOAN$^{*}$~\cite{huang2021toan}     & {66.10} {\scriptsize $\pm$ 0.86}  & {82.27} {\scriptsize $\pm$ 0.60}  & {49.77} {\scriptsize $\pm$ 0.86}  & {69.29} {\scriptsize $\pm$ 0.70}  & {75.28} {\scriptsize $\pm$ 0.72} & {87.45} {\scriptsize $\pm$ 0.48}\\
			\specialrule{0em}{1pt}{1pt}
			BSNet$^{*}$~\cite{li2020bsnet}       & {73.48} {\scriptsize $\pm$ 0.92}  & {83.84} {\scriptsize $\pm$ 0.59} & {61.95} {\scriptsize $\pm$ 0.97} & {79.62} {\scriptsize $\pm$ 0.63} & {71.07} {\scriptsize $\pm$ 1.03} & {88.38} {\scriptsize $\pm$ 0.62}\\
			\specialrule{0em}{1pt}{1pt}
			OLSA$^{*}$~\cite{wu2021object}  & {77.77} {\scriptsize $\pm$ 0.44} & {89.87} {\scriptsize $\pm$ 0.24} & {64.15} {\scriptsize $\pm$ 0.49} & {78.28} {\scriptsize $\pm$ 0.32} & {77.03} {\scriptsize $\pm$ 0.46} & {88.85} {\scriptsize $\pm$ 0.46} \\
			\specialrule{0em}{1pt}{1pt}
			AGPF~\cite{TANG2022108792} & {78.73} {\scriptsize $\pm$ 0.84} & {89.77} {\scriptsize $\pm$ 0.47}& \bf {72.34 {\scriptsize $\pm$ 0.86}} & \bf {84.02 {\scriptsize $\pm$ 0.57}} & {85.34} {\scriptsize $\pm$ 0.74} & {94.79} {\scriptsize $\pm$ 0.35} \\
			\midrule
			\specialrule{0em}{1pt}{1pt}
			Ours    &   \bf {82.27 {\scriptsize $\pm$ 0.46}} & \bf {90.76 {\scriptsize  $\pm$ 0.26}} & {69.58} {\scriptsize  $\pm$ 0.50} & {82.59} {\scriptsize  $\pm$ 0.33} & \bf{88.93 {\scriptsize  $\pm$ 0.38}} & \bf {95.20 {\scriptsize  $\pm$ 0.20}} \\ \bottomrule
	\end{tabular}}	
	\vspace{-4mm}
\end{table*}

\section{Experiments}
\label{set4}

\subsection{Datasets}
\label{train-detail}
We conduct experiments on three widely used benchmarks for FG-FSR, including CUB-200-2011~\cite{wah2011caltech}, Stanford Dogs~\cite{khosla2011novel}, and Stanford Cars~\cite{krause20133d}. We follow the same evaluation protocol as the previous work~\cite{wu2021object,li2019revisiting,TANG2022108792}, and all the input images are resized to $84 \times 84$.
\begin{itemize}
		\item {\bf CUB-200-2011} contains $200$ species of birds with
$11,788$ images in total~\cite{wah2011caltech}. Following the evaluation protocol of~\cite{wu2021object}, we randomly split the dataset into $100$ base categories for training, $50$ categories for validation, and $50$ novel categories for evaluation.
		\item {\bf Stanford Dogs} contains $20,580$ images from $120$ sub-classes of dogs in total~\cite{khosla2011novel}. Following the evaluation protocol of~\cite{li2019revisiting}, we randomly split the dataset into $70$ base categories for training, $20$ categories for validation, and $30$ novel categories for evaluation.
		\item {\bf Stanford Cars} contains $16,185$ images spanning $196$ car models in total~\cite{krause20133d}. Following the evaluation protocol of~\cite{li2019revisiting}, we randomly split the dataset into $130$ base categories for training, $17$ categories for validation, and $49$ novel categories for evaluation.
\end{itemize}

Furthermore, considering that prior work~\cite{zhang2020deepemd,kang2021relational,liu2021dmn4,wertheimer2021few} on CUB-200-2011 pre-processes images with human-annotated bounding box as the input, we also conduct experiments under this setting for a fair comparison.

\begin{table}[tb!]
	\centering
	\caption{\label{tab:2}The accuracy (\%) of $5$-way $1$-shot and $5$-shot tasks on the cropped CUB-$200$-$2011$, where all images are pre-processed with the provided human-annotated bounding boxes. We report  the mean accuracy with 95\% confidence intervals on $2000$ test episodes. Note that ``BB.'' denotes whether to leverage the provided human-annotated bounding boxes to crop the images. $\circ$ denotes the results are reported by~\cite{zhang2020deepemd}. * denotes the results obtained from the original paper.}
	\vspace{-2mm}
    \resizebox{0.9\linewidth}{!}{ 
		\begin{tabular}{lcccc}
			\toprule
			\multirow{2}{*}{\bf Method} & \multirow{2}{*}{\bf BB.}  & \multirow{2}{*}{\bf Backbone} & \multicolumn{2}{c}{\bf CUB-200-2011} \\ \cline{4-5} 
			\specialrule{0em}{1pt}{1pt}
			& & & {\bf 5-way 1-shot} & {\bf 5-way 5-shot}  \\
			\midrule
			\specialrule{0em}{1pt}{1pt}
			MatchingNet$^\circ$~\cite{vinyals2016matching} &\cmark   & ResNet-$12$ & {71.87} {\scriptsize $\pm$ 0.85}& {85.08} {\scriptsize $\pm$ 0.57} \\
			\specialrule{0em}{1pt}{1pt}
			ProtoNet\cite{snell2017prototypical}&\cmark&   ResNet-$12$& {76.21} {\scriptsize $\pm$ 0.48} & {89.70} {\scriptsize $\pm$ 0.27}     \\
			\specialrule{0em}{1pt}{1pt}
			RelationNet~\cite{sung2018learning} &\cmark& ResNet-$12$& {75.81} {\scriptsize $\pm$ 0.53}      & {88.63} {\scriptsize $\pm$ 0.29}    \\
			\specialrule{0em}{1pt}{1pt}
			Baseline~\cite{chen2019closerfewshot} &\cmark& ResNet-$12$& {70.93} {\scriptsize $\pm$ 0.49}      & {88.57} {\scriptsize $\pm$ 0.28}    \\
			\specialrule{0em}{1pt}{1pt}
			Baseline++~\cite{chen2019closerfewshot} &\cmark& ResNet-$12$& {75.17} {\scriptsize $\pm$ 0.49}      & {85.58} {\scriptsize $\pm$ 0.34}    \\
			\specialrule{0em}{1pt}{1pt}
			DeepEMD$^{*}$~\cite{zhang2020deepemd} &\cmark & ResNet-$12$ & {75.65} {\scriptsize $\pm$ 0.83} & {88.69} {\scriptsize $\pm$ 0.50}    \\
			\specialrule{0em}{1pt}{1pt}
			RENet$^{*}$~\cite{kang2021relational}&\cmark & ResNet-$12$& {79.49} {\scriptsize $\pm$ 0.44} & {91.11} {\scriptsize $\pm$ 0.24}     \\ 
			\specialrule{0em}{1pt}{1pt}
			DMN4$^{*}$~\cite{liu2021dmn4}  &\cmark & ResNet-$12$& {82.95} {\scriptsize $\pm$ 0.75}& {90.46}  {\scriptsize $\pm$ 0.46} \\ 
			\specialrule{0em}{1pt}{1pt}
			FRN~\cite{wertheimer2021few}  &\cmark & ResNet-$12$& {82.12} {\scriptsize $\pm$ 0.85} & {92.49} {\scriptsize $\pm$ 0.43}  \\ 
			\midrule
			\specialrule{0em}{1pt}{1pt}
			Ours & \xmark & ResNet-$12$  & {82.27} {\scriptsize $\pm$ 0.46}&{90.76}  {\scriptsize $\pm$ 0.26}\\
			\specialrule{0em}{1pt}{1pt}
			Ours & \cmark & ResNet-$12$  & \bf {86.00 {\scriptsize  $\pm$ 0.41}}& \bf {92.53} {\scriptsize $\pm$ 0.23}\\ \bottomrule
	\end{tabular}}
	\vspace{-4mm}
\end{table}

\subsection{Implementation Details}

\begin{table*}[t!]
	\centering
	\caption{ \label{tab:abB&L} Comparison results on backbones with/without two-stage local feature-based similarity metric. \texttt{LOCAL}: Removing GAP and using local features. \texttt{Raw}: Employing features from raw images. \texttt{Refined}: Employing features from refined images.}
	\vspace{-2mm}
	\resizebox{0.9\linewidth}{!}{ 
		\begin{tabular}{lccccccccc}
			\toprule
			\specialrule{0em}{1pt}{1pt}
			\multirow{2}{*}{\bf Model} &\multirow{2}{*}{\texttt{LOCAL}}  &\multirow{2}{*}{\texttt{RAW}} &\multirow{2}{*}{\texttt{REFINED}} &\multicolumn{2}{c}{\bf CUB-200-2011} & \multicolumn{2}{c}{\bf Stanford Dogs} & \multicolumn{2}{c}{\bf Stanford Cars} \\ 
			\cline{5-10}
			\specialrule{0em}{1pt}{1pt}
			&&&&{$5$-way $1$-shot} & {$5$-way $5$-shot}& {$5$-way $1$-shot} &{$5$-way $5$-shot}& {$5$-way $1$-shot} &{$5$-way  $5$-shot}  \\
			\midrule
			\specialrule{0em}{1pt}{1pt}
			B0  &&\checkmark&& {77.44} {\scriptsize $\pm$ 0.50} & {86.82} {\scriptsize $\pm$ 0.32} & {65.80} {\scriptsize $\pm$ 0.51}& {79.66} {\scriptsize $\pm$ 0.36} & {85.20} {\scriptsize $\pm$ 0.44}& {92.87} {\scriptsize $\pm$ 0.24}\\
			\specialrule{0em}{1pt}{1pt}
			B1  &\checkmark&\checkmark&& {78.75} {\scriptsize $\pm$ 0.49}  & {88.96} {\scriptsize $\pm$ 0.28} & {67.18} {\scriptsize $\pm$ 0.50}     & {81.27} {\scriptsize $\pm$ 0.34} & {85.87} {\scriptsize $\pm$ 0.43} & {93.85} {\scriptsize $\pm$ 0.24} \\
			\specialrule{0em}{1pt}{1pt}
			  B2 &\checkmark&&\checkmark& {78.35}  {\scriptsize $\pm$ 0.50} & {87.62} {\scriptsize $\pm$ 0.30} &  {64.56} {\scriptsize $\pm$ 0.51}  & {78.24} {\scriptsize $\pm$ 0.35}  & {84.04} {\scriptsize $\pm$ 0.45} & {92.50} {\scriptsize $\pm$ 0.25}\\
			\specialrule{0em}{1pt}{1pt}
			B3 &\checkmark&\checkmark&\checkmark& \bf{80.69}  {\scriptsize $\pm$ 0.48} & \bf{90.21} {\scriptsize $\pm$ 0.27} &  \bf{68.89} {\scriptsize $\pm$ 0.49}  & \bf{82.06} {\scriptsize $\pm$ 0.33}  & \bf{87.65} {\scriptsize $\pm$ 0.40} & \bf{94.85} {\scriptsize $\pm$ 0.21}\\ \bottomrule
	\end{tabular}}	
	\vspace{-2mm}
\end{table*}

\begin{table*}[tb!]
	\centering
	\caption{ \label{tab:abBFL} Contribution of each individual component in our method. \texttt{BAS}: background activation suppression. \texttt{L2L}: Local to local similarity metric. \texttt{FOA}: Foreground object alignment. \texttt{AE}: Attentive erasing.}
	\vspace{-2mm}
	\resizebox{0.9\linewidth}{!}{ 
		\begin{tabular}{lcccccccccc}
			\toprule
			\specialrule{0em}{1pt}{1pt}
			\multirow{2}{*}{ \bf Model} & \multirow{2}{*}{\texttt{BAS} } & \multirow{2}{*}{\texttt{L2L}} & \multirow{2}{*}{\texttt{FOA}} & \multirow{2}{*}{\texttt{AE}} &\multicolumn{2}{c}{\bf CUB-200-2011} & \multicolumn{2}{c}{\bf Stanford Dogs} & \multicolumn{2}{c}{\bf Stanford Cars} \\ 
			\cline{6-11}
			\specialrule{0em}{1pt}{1pt}
			&&&&&{$5$-way $1$-shot} & {$5$-way $5$-shot}& {$5$-way $1$-shot} &{$5$-way $5$-shot}& {$5$-way $1$-shot} &{$5$-way  $5$-shot}  \\
			\midrule
			\specialrule{0em}{1pt}{1pt}
			C0  &\checkmark&&&& {78.18} {\scriptsize $\pm$ 0.50} & {87.28} {\scriptsize $\pm$ 0.31} & {66.85} {\scriptsize $\pm$ 0.52}& {79.68} {\scriptsize $\pm$ 0.36} & {86.03} {\scriptsize $\pm$ 0.42}& {92.81} {\scriptsize $\pm$ 0.25}\\
			\specialrule{0em}{1pt}{1pt}
			C1  &\checkmark&\checkmark&&& {80.69} {\scriptsize $\pm$ 0.48}  & {90.21} {\scriptsize $\pm$ 0.27} & {68.89} {\scriptsize $\pm$ 0.49}     & {82.06} {\scriptsize $\pm$ 0.33} & {87.65} {\scriptsize $\pm$ 0.40} & {94.85} {\scriptsize $\pm$ 0.21} \\
			\specialrule{0em}{1pt}{1pt}
			C2  &\checkmark&\checkmark&\checkmark&& {81.71}  {\scriptsize $\pm$ 0.47} & {90.49} {\scriptsize $\pm$ 0.27} &  {69.09} {\scriptsize $\pm$ 0.50}  & {82.25} {\scriptsize $\pm$ 0.33}  & {88.93} {\scriptsize $\pm$ 0.39} & {95.00} {\scriptsize $\pm$ 0.20}
			\\
			\specialrule{0em}{1pt}{1pt}
			C3 &\checkmark&\checkmark&\checkmark&\checkmark& \bf{82.27}  {\scriptsize $\pm$ 0.46} & \bf{90.76} {\scriptsize $\pm$ 0.26} &  \bf{69.58} {\scriptsize $\pm$ 0.50}  & \bf{82.59} {\scriptsize $\pm$ 0.33}  & \bf{88.93} {\scriptsize $\pm$ 0.38} & \bf{95.20} {\scriptsize $\pm$ 0.20}
			\\ 
				\specialrule{0em}{1pt}{1pt}
			C4 &&\checkmark&\checkmark&\checkmark& {79.75}  {\scriptsize $\pm$ 0.48} & {89.44} {\scriptsize $\pm$ 0.28} &  {67.80} {\scriptsize $\pm$ 0.51}  & {81.59} {\scriptsize $\pm$ 0.33}  & {88.19} {\scriptsize $\pm$ 0.39} & {94.37} {\scriptsize $\pm$ 0.22}
			\\ 
			\bottomrule
	\end{tabular}}	
	\vspace{-2mm}
\end{table*}

To have a fair and broad comparison with the previous work~\cite{wu2021object,zhang2020deepemd,li2020bsnet,huang2021toan}, we adopt the widely-used backbone (\ie,~ResNet-$12$) as the feature extractor. 
ResNet-$12$ consists of $4$ consecutive basic blocks and the number of filters is set to $64$-$128$-$256$-$512$. 
To achieve more accurate background suppression and foreground alignment, we remove the last pooling layer of ResNet-$12$, thus the size of the output feature map is $512\times 11\times 11$.
Without bells and whistles, no additional parameters are introduced in our model except two fully connected (FC) layers are used as global classifiers. 

During the training stage, standard data augmentation strategies are adopted, \eg, random cropping and random horizontal flipping.
For all experiments, we utilize SGD with an initial learning rate of $0.1$ as the optimizer to train our model on the Titan RTX GPU for $90$ epochs. The learning rate decays to $0.06$ in the $60{\text{th}}$ epoch and then times $0.2$ every $10$ epochs. 
The $\lambda$ in the overall loss is experimentally set to $0.1$.
Significantly, our proposed framework is trained from scratch in an end-to-end manner, and there is no need to pre-train the feature extractor on other datasets. During the testing stage, evaluation is performed under standard `$5$-way $1$-shot' and `$5$-way $5$-shot' settings, we test $2,000$ episodes and report the mean accuracy with $95\%$ confidence intervals as the final result. Our experiments are implemented in PyTorch.

\subsection{Comparison With State-of-the-Art Methods}
\subsubsection{\bf{Results on general fine-grained datasets}}
Tab.~\ref{tab:1} shows the performance evaluations on three aforementioned fine-grained benchmark datasets. We demonstrate ten representative methods that have reported evaluation results on the corresponding datasets, including six typical FSL methods (\ie,~ProtoNet~\cite{snell2017prototypical}, RelationNet~\cite{sung2018learning}, DN4~\cite{li2019revisiting}, Baseline~\cite{chen2019closerfewshot}, Baseline++~\cite{chen2019closerfewshot}, and CAN~\cite{hou2019cross}) and four specialized FS-FGR methods (\ie,~TOAN~\cite{huang2021toan}, BSNet~\cite{li2020bsnet}, OLSA~\cite{wu2021object}, and AGPF~\cite{TANG2022108792}).
For a fair comparison with the above methods, we display the results of our proposed method based on the ResNet-$12$ backbone. Compared with the above typical FSL and FS-FGR methods, our proposed method achieves the highest performance under both `$5$-way $1$-shot' and `$5$-way $5$-shot' settings on the most benchmark datasets. The observations can be summarized as follows:
\begin{itemize}
		\item On the CUB-200-2011 dataset, our proposed method achieves stable and excellent performance and surpasses all existing methods by a large margin in terms of `$5$-way $1$-shot' setting. It's clear that our method exceeds six typical FSL methods (\ie~\cite{snell2017prototypical,sung2018learning,li2019revisiting,chen2019closerfewshot,hou2019cross}) significantly. Compared with the global feature based FS-FGR method TOAN~\cite{huang2021toan}, our proposed method achieves $16.17\%$ and $8.49\%$ performance improvement under $1$-shot and $5$-shot settings, respectively. In addition, the current state-of-the-art accuracy of $1$-shot is achieved by AGPF~\cite{TANG2022108792} with $78.73\%$, which shares the same insights with our method. Our method outperforms AGPF by $3.54\%$ with an accuracy of $82.27\%$.
		\item On the Stanford Dogs dataset, our method overall exceeds all typical FSL methods significantly in terms of $1$-shot and $5$-shot settings. But compared with AGPF~\cite{TANG2022108792}, our method is lower than it by $2.76\%$ and $1.43\%$ under $1$-shot and $5$-shot settings, respectively. Note that the AGPF model currently gets the highest performance, which mainly benefits from additional structures used to construct the feature pyramid and attention pyramid. In contrast, the operations of background suppression and foreground alignment in our method are \textit{parameter-free}, which has significant advantages in efficiency.
		\item On the Stanford Cars dataset, our method again outperforms all other compared methods with the best performance of $88.93\%$ for $1$-shot and $95.20\%$ for $5$-shot. Compared to the leading results obtained by AGPF~\cite{TANG2022108792} and CAN~\cite{hou2019cross}, the relative performance
        gains are $2.03$\% and $0.41$\% for $1$-shot and $5$-shot, respectively. It's worth noting that cars are rigid objects and the corresponding intra-category variances are not as significant as birds and dogs. Our methods can reach the best performance of $95.20\%$ under $5$-shot setting, which further confirms the significance of our method for FG-FSR.
\end{itemize}

\subsubsection{\bf{Results on CUB-200-2011 with bounding boxes}}
Recently, some state-of-the-art FSL methods (\eg~DeepEMD ~\cite{zhang2020deepemd}, RENet~\cite{kang2021relational}, DMN4~\cite{liu2021dmn4}, \textit{etc}) also demonstrate their generalization ability on the fine-grained dataset, where the images from CUB-$200$-$2011$ are pre-processed by human-annotated bounding boxes. In this section, we also carry out extra experiments to compare our method with these state-of-the-art methods under the same benchmark (\ie, cropped CUB-$200$-$2011$) for a fair comparison. 
The cropped CUB-$200$-$2011$ is less challenging because object images cropped by human-annotated bounding boxes have a finer scale and less background, which is helpful to study the fine-grained characteristics for recognition. 
As shown in Tab.~\ref{tab:2}, our method overall exceeds all compared methods in terms of $1$-shot and $5$-shot setting and establishes a novel state-of-the-art. 
Significantly, it can be observed that:
\begin{itemize}
\item
Compared with the current state-of-the-art methods trained on the cropped CUB-$200$-$2011$, our method trained on the uncropped CUB-$200$-$2011$ (\ie,~w/o BB.) can achieve competitive performance. It is well-known that RENet~\cite{kang2021relational} and DMN4~\cite{liu2021dmn4} are the two latest methods in the FSL community. Our method achieves excellent results in the case of not using the cropped dataset, even $2.78\%$ higher than RENet~\cite{kang2021relational} under $1$-shot setting and $0.30\%$ higher than DMN4~\cite{liu2021dmn4} under $5$-shot setting. In fact, cropping object images with human-annotated bounding boxes is equivalent to perfectly removing background noises. This proves the great contribution of \textit{background suppression} to improve the performance of FS-FGR, which also proves the advantages of our method.
\item
When introducing an additional bounding box to pre-process the CUB-$200$-$2011$, our method surpasses all existing methods by a large margin, especially under the $1$-shot setting. What's more, compared with the results obtained from the uncropped dataset, our method (\ie,~w/~BB.) improves  $3.73\%$ and $1.77\%$ in terms of $1$-shot and $5$-shot setting, respectively. For the cropped CUB-200-2011, the performance of our method is further improved significantly without the interference of cluttered background. In this case, the excellent performance mainly benefits from two aspects: (1) BAS can further enable the model to crop out the key region of the image with the most discrimination and less redundancy; (2) The \textit{foreground alignment} operation of FOA can enable the model to learn fine-grained features of different parts with different importance degree.
Note that the FRN~\cite{wertheimer2021few} model currently gets the state-of-the-art performance for FS-FGR. Our method, however, still surpasses FRN by $3.88\%$ under the $1$-shot setting, which further demonstrates that the proposed method has a significant superiority under the $1$-shot setting. 
\end{itemize}

\subsection{Ablation Studies}
\label{ab}
In this section, we conduct ablation studies to investigate the effectiveness of each individual component of our method. The following experiments are all conducted on the three fine-grained datasets with ResNet-$12$ as the backbone if not particularly mentioned. We first build a group of baseline models as follows:

\begin{itemize}
    \item {\bf B0: Global features for FS-FGR} refers to employing backbone to extract global features from raw images, which are directly used to compute the \textit{cosine distances} between the query images and support prototypes. This aims to test the basic performance of our full model by removing BAS, FOA, and L2L. Here, we also retain an auxiliary  global classifier to learn in the whole base category space.
    \item {\bf B1: Raw Local features for FS-FGR} refers to employing the backbone to extract local features only from raw images, which are used to compute the similarity between the query images and support prototypes. This aims to test the superiority of raw local features by adding L2L on the basis of \textbf{B0}.
    \item {\bf B2: Refined Local features for FS-FGR} refers to employing the backbone to extract local features only from refined images, which are used to compute the similarity between the query images and support prototypes. This aims to test the superiority of refined local features by adding L2L and BAS on the basis of \textbf{B0}.
    \item {\bf B3: Dual Local features for FS-FGR} refers to employing backbone to extract local features from raw images and refined images, where both of them are used to compute the similarity between the query images and support prototypes. This aims to test the superiority of our two-stage workflow by adding L2L and BAS on the basis of \textbf{B0}.
\end{itemize}

\begin{table}[t!]
	\centering
	\caption{ \label{tab:conv4}The accuracy (\%) of $5$-way $1$-shot and $5$-shot tasks on CUB-200-2011 with various backbones (\ie,~Conv-$64$ and ResNet-$18$). The results are obtained from the original paper. We report the mean accuracy with 95\% confidence intervals on $2000$ test episodes. The best results are shown in bold.}
	\resizebox{0.9\linewidth}{!}{ 
		\begin{tabular}{lccc}
			\toprule
			\specialrule{0em}{1pt}{1pt}
\multirow{2}{*}{\bf Method} & \multirow{2}{*}{\bf Backbone} &\multicolumn{2}{c}{\bf CUB-200-2011} \\
			\cline{3-4}
			\specialrule{0em}{1pt}{1pt}
			& & {$5$-way $1$-shot} & {$5$-way $5$-shot}  \\
			\midrule
			\specialrule{0em}{1pt}{1pt}
			DN$4$~\cite{li2019revisiting}  & Conv-$64$& {53.15} {\scriptsize $\pm$ 0.84} & \bf{81.90} {\scriptsize $\pm$ 0.60} \\
			\specialrule{0em}{1pt}{1pt}
			MattML~\cite{zhu2020multi}  & Conv-$64$ & \bf{66.29} {\scriptsize $\pm$ 0.56}  & {80.34} {\scriptsize $\pm$ 0.30}  \\
			\specialrule{0em}{1pt}{1pt}
			BSNet~\cite{li2020bsnet} & Conv-$64$ & {65.89}  {\scriptsize $\pm$ 1.00} & {80.99} {\scriptsize $\pm$ 0.63}  \\
			\specialrule{0em}{1pt}{1pt}
			TOAN~\cite{huang2021toan}  & Conv-$64$ & {65.34}  {\scriptsize $\pm$ 0.75} & {80.43} {\scriptsize $\pm$ 0.60} \\
			\specialrule{0em}{1pt}{1pt}
			Ours  & Conv-$64$& {65.48} {\scriptsize $\pm$ 0.51} & {76.02} {\scriptsize $\pm$ 0.41} \\
			\hline
			\specialrule{0em}{1pt}{1pt}	ProtoNet~\cite{snell2017prototypical}  & ResNet-$18$& {72.99} {\scriptsize $\pm$ 0.88} & {86.64} {\scriptsize $\pm$ 0.51} \\
			\specialrule{0em}{1pt}{1pt}	
			Neg-Margin~\cite{LiuCLL0LH20}  & ResNet-$18$& {72.66} {\scriptsize $\pm$ 0.85} & {89.40} {\scriptsize $\pm$ 0.43} \\
			\specialrule{0em}{1pt}{1pt}	
			Centroid~\cite{AfrasiyabiLG20}  & ResNet-$18$& {74.22} {\scriptsize $\pm$ 1.09} & {88.65} {\scriptsize $\pm$ 0.55} \\
			\specialrule{0em}{1pt}{1pt}	
			RAP-Neg-Margin~\cite{HongFL0SHP21}  & ResNet-$18$& {75.37} {\scriptsize $\pm$ 0.81} & {90.61} {\scriptsize $\pm$ 0.39} \\
			\specialrule{0em}{1pt}{1pt}
			Ours & ResNet-$18$ & \bf{84.19}  {\scriptsize $\pm$ 0.43} & \bf{92.46} {\scriptsize $\pm$ 0.23} \\ \bottomrule
	\end{tabular}}	
	\vspace{-4mm}
\end{table}

The comparison results among the above baselines are shown in Table~\ref{tab:abB&L}. In this part, we first investigate the significance of the local feature-based similarity metric for FS-FGR. Obviously, B1-B3 (using local features) outperform B0 (using global features) and lead to consistent performance gains on three datasets. This validates that using local features to compute image-to-image similarity can avoid losing subtle information.
With the proposed L2L, we have observed that the performance of B2 is a little lower than B1. That is to say, the baseline model using only the refined image obtained by BAS is inferior to the model using the raw image on each benchmark. 
The underlying reason could be that the low resolution (\ie,~$84\times84$) of the input image results in inaccurate localization of the completed region for the foreground as shown in Fig.~\ref{fig:vis}. In fact, although BAS can highlight foreground objects, it also leads to some suppression results on some parts of objects, due to the imbalanced feature distribution between foreground and background.
However, B3 integrated with local features from both the raw images and refined images achieves the best performance, this verifies the necessity and effectiveness of background suppression for FS-FGR. In addition, a substantial improvement achieved by B3 reveals that our method equipped with BAS considers both the raw image and refined image to alleviate the problem of insufficient samples and makes the model robust to varieties of object scales.

\begin{table}[t!]
 \scriptsize
	\centering
	\caption{\label{tab:com}Comparison of the efficiency of some publicly available local feature based few-shot learning methods. Smaller Params. and smaller FLOPs indicate better efficiency.}
	\resizebox{0.9\linewidth}{!}{ 
		\begin{tabular}{lccc}
			\toprule
			\specialrule{0em}{1pt}{1pt}
\multirow{2}{*}{\bf Method} & \multirow{2}{*}{\bf Backbone} &\multicolumn{2}{c}{\bf Model Complexity} \\
			\cline{3-4}
			\specialrule{0em}{1pt}{1pt}
			& & {Params.~(M)} & {FLOPs~(G)}  \\
			\midrule
			\specialrule{0em}{1pt}{1pt}
			DN4~\cite{li2019revisiting}  & ResNet-$12$ & 12.42 & 67.39  \\
			\specialrule{0em}{1pt}{1pt}
			CAN~\cite{hou2019cross}  & ResNet-$12$ & 8.04 & 12.75  \\
			\specialrule{0em}{1pt}{1pt}
			DeepEMD~\cite{zhang2020deepemd}  & ResNet-$12$ & 12.42 & 35.23  \\
			\specialrule{0em}{1pt}{1pt}
			RENet~\cite{kang2021relational}  & ResNet-$12$ & 12.63 & 35.70 \\
			\specialrule{0em}{1pt}{1pt}
			DMN4~\cite{liu2021dmn4}  & ResNet-$12$ & 12.42 & 35.23 \\
			\specialrule{0em}{1pt}{1pt}
			FRN~\cite{wertheimer2021few}  & ResNet-$12$ &12.42 & 35.18 \\
			\specialrule{0em}{1pt}{1pt}
			AGPF~\cite{TANG2022108792}  & ResNet-$12$ & 8.77& 51.53 \\
			\midrule
			\specialrule{0em}{1pt}{1pt}
			Ours (1-stage)  & ResNet-$12$& 8.04 & 16.88
			\\ 
			\specialrule{0em}{1pt}{1pt}
			Ours (2-stage)  & ResNet-$12$& 8.04 & 33.76
			\\ 
			\bottomrule
	\end{tabular}}	
	\vspace{-4mm}
\end{table}

As mentioned above, BAS, L2L, and FOA are three key components of our proposed method. Thus, we also conduct the ablation experiments to analyze the contribution of each component by setting four baselines, as follows:
\begin{itemize}
    \item {\bf C0: Background suppression for FS-FGR} refers to employing BAS on the backbone to extract global features from raw images and refind images. Both of images are used to compute the \textit{cosine distances} between the query images and support prototypes. This aims to test the superiority of background suppression by adding BAS based on \textbf{B0}.
    \item {\bf C1: Local to local similarity metric for FS-FGR} refers to employing the backbone to extract local features from raw and refined images. Both images are used to compute the similarity between the query images and support prototypes. This aims to test the superiority of the local feature-based similarity metric by adding L2L based on \textbf{C0}.
    \item {\bf C2: Foreground alignment for FS-FGR} refers to employing the backbone to first extract local features from raw images and refined images and then align them. Both images are used to compute the similarity between the query images and support prototypes. This aims to test the superiority of aligned local features by adding FOA based on \textbf{C1}.
    \item {\bf C3: Attentive erasing for FS-FGR} refers to erasing the region with activation higher than a threshold to further improve the completeness of foreground based on BAS.  This aims to test the superiority of erasing strategy in our method by adding attentive erasing to the basis of \textbf{C2}.
    \item {\bf C4: Foreground alignment w/o background suppression for FS-FGR} refers to only employing backbone to extract local features from raw images and then align them to compute the similarity between the query images and support prototypes. This aims to verify the complementarity of background suppression and foreground alignment by removing BAS based on \textbf{C3}.
\end{itemize}

\begin{figure}[t!]
    \centering
    \includegraphics[scale=0.3]{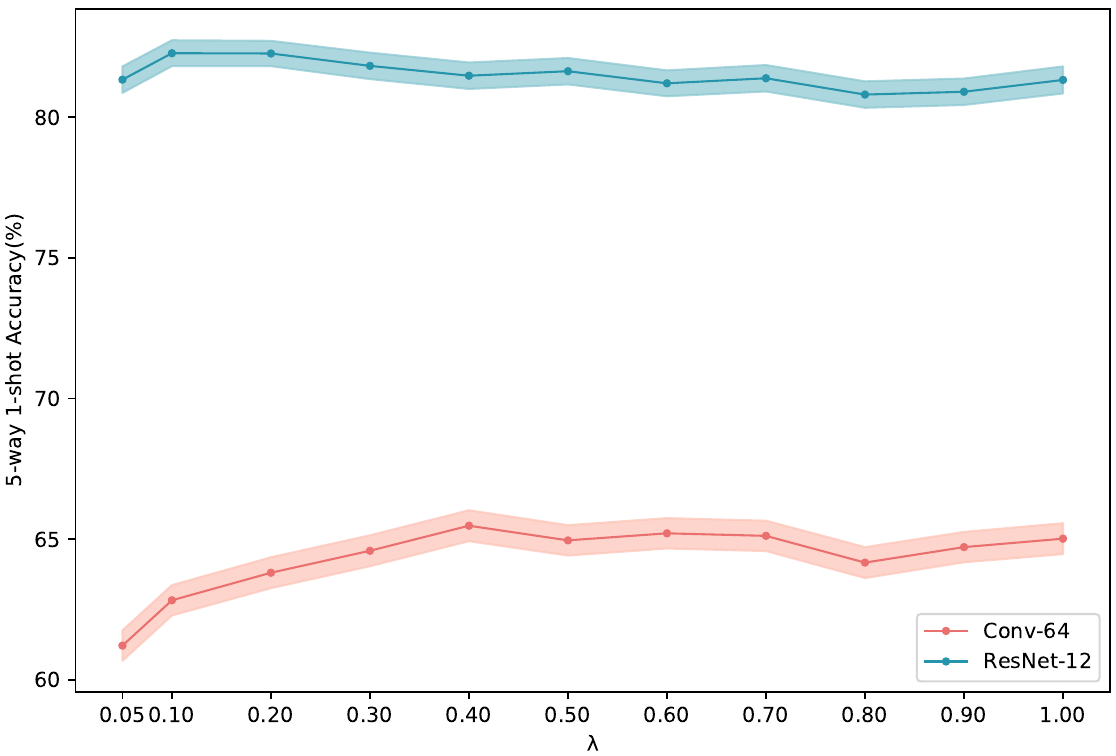}
    \vspace{-2mm}
    \caption{Performance with respect to different values of $\lambda$ on CUB-200-2011 under 5-way 1-shot setting. The blue line shows the results of ResNet-$12$, and the red line shows the results of Conv-$64$.}
    \label{fig:lambda}
    	\vspace{-4mm}
\end{figure}

Table~\ref{tab:abBFL} lists the comparison results of the ablation study when using our proposed modules. We have the following observations. First, the addition of BAS solely can only contribute to limited improvement in performance when using global features to compute similarity (\ie, C0 \textit{vs.} B0). Second, when introducing local features to measure the distances between the query images and support prototypes, L2L significantly improves the localization accuracy under $1$-shot and $5$-shot settings (\ie, C1 \textit{vs.} C0). Third, a substantial improvement of performance on each dataset is gained by integrating FOA to align support features and query features (\ie, C2 \textit{vs.} C1). This verifies the necessity of considering variations in the object's pose and position for the FS-FGR task. Finally, erasing the regions with activation higher than a threshold during the training phase alleviates the problem of the model focusing on the most discriminative regions (\ie, C3 \textit{vs.} C2). 
The aforementioned experiments and comparisons have demonstrated that the addition of FOA based on BAS can significantly improve the performance of FS-FGR. Moreover, we further demonstrate that BAS and FOA are complementary to each other in improving performance by removing BAS from the full model (\ie, C4 \textit{vs.} C3). Meanwhile, we observe that C4 outperforms B1 by a large margin, demonstrating the superiority of our feature alignment in the absence of background suppression.

\subsection{Model Complexity Analysis}
To comprehensively assess the performance, we compare the efficiency of our method with some publicly available local feature based few-shot learning methods. The results of model parameters (Params.) and floating-point operations (FLOPs) are reported in Table~\ref{tab:com}. From the aforementioned experiments and comparisons in Table~\ref{tab:1} and Table~\ref{tab:2}, although the proposed method adopts a two-stage framework that would double the amount of computation and storage, we can see that our method still achieves better FS-FGR performance with competitive computational efficiency. Our proposed background activation suppression operation can be theoretically conducted on any position of the network by generating a class-agnostic activation map for localization. With the consideration of both accuracy and efficiency, we consequently choose the high-level features from the last layer of the feature extraction. In summary, without bells and whistles, the efficiency of the proposed two-stage method can benefit from three aspects: (1) the background suppression and foreground alignment operations are only directly conducted on the feature maps without introducing additional parameters; (2) the raw stage and the refined stage are based on the same network with shared parameters, which limits the increase of the number of model parameters; (3) the background suppression and foreground alignment are conducted on high-level features with minimum resolutions.

\begin{table}[tb!]
	\centering
	\caption{ \label{tab:alpha_beta}The accuracy (\%) comparison of $5$-way $1$-shot and $5$-shot tasks on CUB-200-2011 by adjusting two weighting factors in the loss functions. We report the mean accuracy with 95\% confidence intervals on $2000$ test episodes.}
	\vspace{-2mm}
	\resizebox{0.9\linewidth}{!}{ 
		\begin{tabular}{ccccc}
			\toprule
			\specialrule{0em}{1pt}{1pt}
\multirow{2}{*}{\bf $\alpha$} &\multirow{2}{*}{\bf $\beta$} & \multirow{2}{*}{\bf Backbone} &\multicolumn{2}{c}{\bf CUB-200-2011} \\
			\cline{4-5}
			\specialrule{0em}{1pt}{1pt}
			 &  & & {$5$-way $1$-shot} & {$5$-way $5$-shot}  \\
			\midrule
			\specialrule{0em}{1pt}{1pt}
			0.1 & 0.9  & Conv-$64$& {61.79} {\scriptsize $\pm$ 0.55} & {76.39} {\scriptsize $\pm$ 0.41} \\
			\specialrule{0em}{1pt}{1pt}
		0.3 & 0.7 & Conv-$64$ & {63.84} {\scriptsize $\pm$ 0.54}  & {75.73} {\scriptsize $\pm$ 0.42}  \\
			\specialrule{0em}{1pt}{1pt}
		\bf{0.5} & \bf{0.5} & Conv-$64$ & \bf{65.48}  {\scriptsize $\pm$ 0.51} & \bf{76.02} {\scriptsize $\pm$ 0.41}  \\
			\specialrule{0em}{1pt}{1pt}
		0.7 & 0.3 & Conv-$64$ & {62.97}  {\scriptsize $\pm$ 0.53} & {75.95} {\scriptsize $\pm$ 0.43} \\
			\specialrule{0em}{1pt}{1pt}
		0.9 & 0.1  & Conv-$64$& {62.02} {\scriptsize $\pm$ 0.53} & {75.62} {\scriptsize $\pm$ 0.43} \\
			\hline
			\specialrule{0em}{1pt}{1pt}	
		0.1 & 0.9 & ResNet-$12$& {81.06} {\scriptsize $\pm$ 0.48} & {89.68} {\scriptsize $\pm$ 0.28} \\
			\specialrule{0em}{1pt}{1pt}	
		0.3 & 0.7 & ResNet-$12$& {81.73} {\scriptsize $\pm$ 0.46} & {90.34} {\scriptsize $\pm$ 0.26} \\
			\specialrule{0em}{1pt}{1pt}	
		\bf{0.5} &\bf{0.5} & ResNet-$12$& \bf{82.27} {\scriptsize $\pm$ 0.46} & \bf{90.76} {\scriptsize $\pm$ 0.26} \\
			\specialrule{0em}{1pt}{1pt}	
		0.7 & 0.3 & ResNet-$12$& {80.76} {\scriptsize $\pm$ 0.48} & {90.16} {\scriptsize $\pm$ 0.27} \\
			\specialrule{0em}{1pt}{1pt}
	   0.9 & 0.1 & ResNet-$12$ & {79.81}  {\scriptsize $\pm$ 0.48} & {89.07} {\scriptsize $\pm$ 0.28} \\ \bottomrule
	\end{tabular}}	
	\vspace{-4mm}
\end{table}

\subsection{Backbones Analysis}
\begin{figure}[t!]
    \centering
    \includegraphics[scale=0.45]{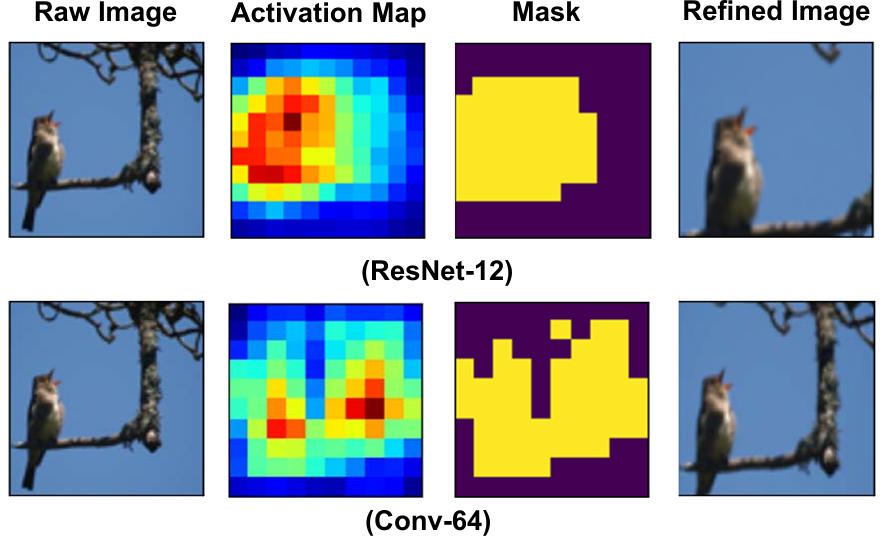}
    \vspace{-3mm}
    \caption{Visualization comparison of background activation suppression with different backbones on CUB-$200$-$2011$ dataset. It can be found that a more compact activation map, complete mask, and accurate refined image can be generated when using deeper ResNet-12 as a feature extractor.}
    \label{fig:vis1}
    \vspace{-4mm}
\end{figure}

In the previous experiments, we used ResNet-$12$ as the backbone to extract local features and compute similarities. To further investigate the effectiveness of the proposed method, we switch to another two embedding backbones from shallow to deep, namely Conv-$64$ and ResNet-$18$, which are commonly used in the FSL community. Compared to ResNet-$18$, Conv-$64$ is shallower with only four layers. Each convolution layer is designed with $3\times3$ convolution of 64 filters, followed by batch normalization, a $\mathbf{RELU}$ activation function, and a $2\times2$ max-pooling. To keep consistent with the previous experimental settings, and to preserve the spatial resolution of features, we remove the last max-pooling layer of Conv-$64$. Thus, the size of the output feature maps becomes $64\times 10\times 10$. 
The top half of Table~\ref{tab:conv4} shows that our method achieves stable and competitive performance under $1$-shot setting when using Conv-$64$. Our method, however, performs worse than some recent methods under the $5$-shot setting. 
The underlying reason could be that Conv-$64$ is too shallow and the poor feature expression ability can not approximate the spatial distribution of the foreground efficiently, which results in the model not being able to generate good foreground object coordinates for background suppression as shown in Fig.~\ref{fig:vis1}. Thus, sub-optimal refined images cannot promote the performance of our method and even bring some negative effects. The existing state-of-the-art FS-FGR models also embed additional complex structures in the basic feature extractor, where redundant parameters can significantly promote the performance of shallow backbones. Compared with them, our model has no redundant structure and parameter except the two global classifiers. Significantly, the operations of background suppression and foreground alignment in our method are \textit{parameter-free}, which has significant advantages in efficiency.
As shown in the bottom half of Table~\ref{tab:conv4}, when using a deeper backbone (\ie, ResNet-$18$, the standard unchanged architecture used in \cite{HeZRS16}) as feature extractor, our method overall exceeds all compared methods and further establishes a novel state-of-the-art. Compared with ResNet-$12$, ResNet-$18$ achieves a significant improvement of $1.92\%$ under a 5-way 1-shot setting and $1.70\%$ under a 5-way 5-shot setting, respectively. The reason may be that the deeper feature extractor can significantly improve the accuracy of background activation suppression and enhance the semantic correlation of foreground object alignment.

\begin{figure}[tb!]
    \centering
    \includegraphics[scale=0.34]{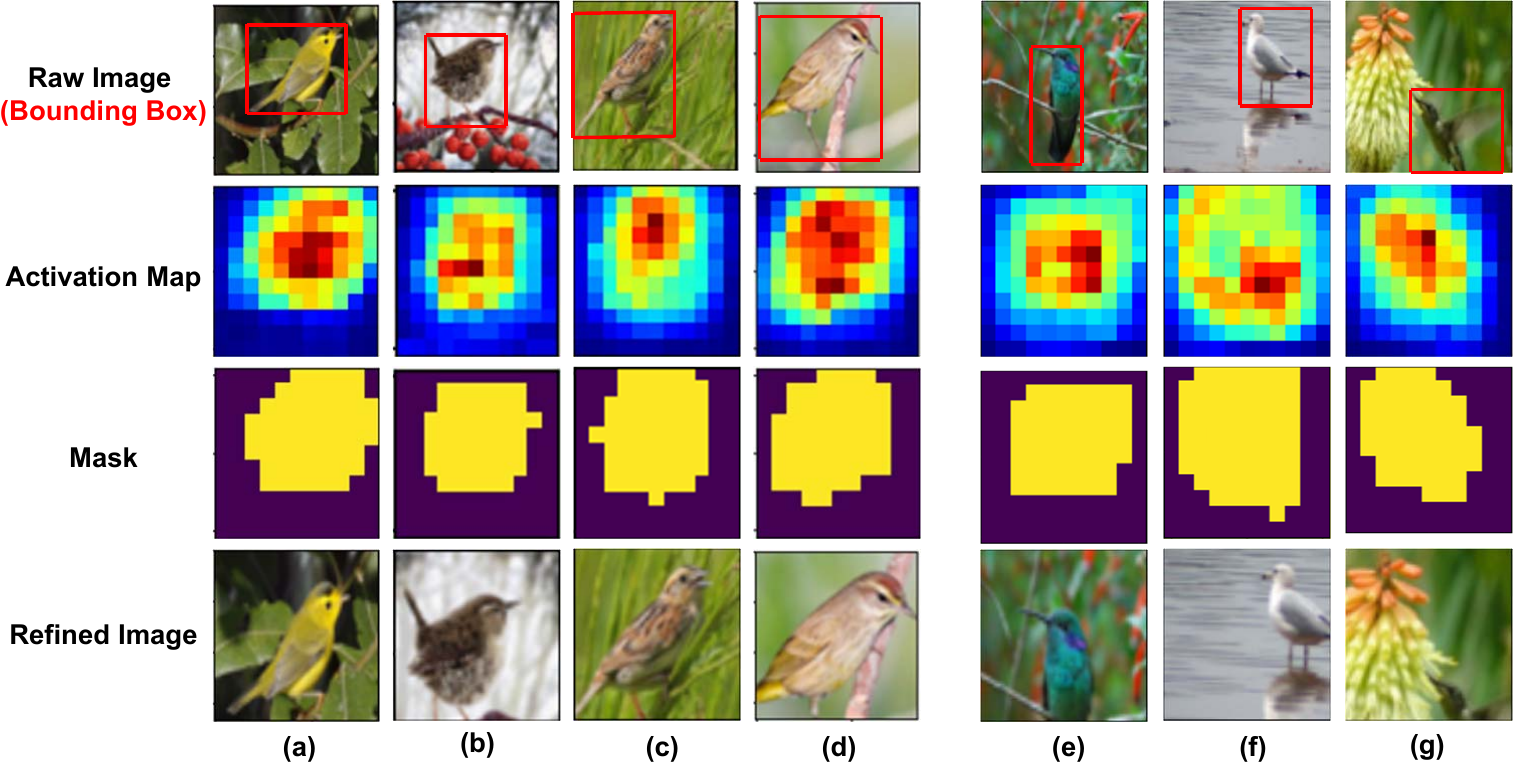}
    \vspace{-2mm}
    \caption{Visualization of background activation suppression under 5-way 1-shot setting on CUB-$200$-$2011$ dataset, including some success and failure cases. For better comparison, the ground-truth bounding box provided by the original dataset is marked in each raw image.}
    \label{fig:vis}
    \vspace{-2mm}
\end{figure}

\begin{figure}[tb!]
    \centering
    \includegraphics[scale=0.3]{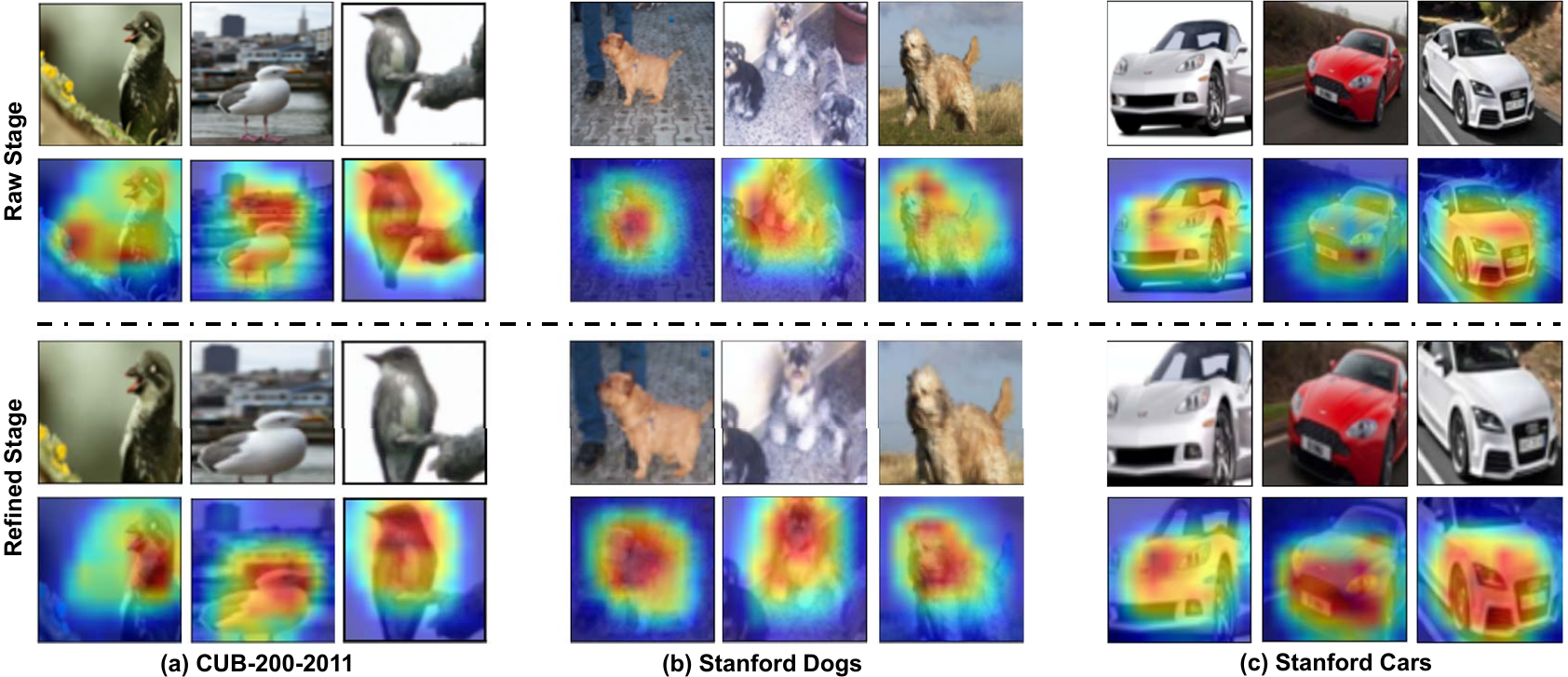}
    \vspace{-2mm}
    \caption{Visualization results of heatmaps in the raw stage and refined stage. We can observe that the fine-grained visual cues of the attended regions from the refined stage are more clear and discriminative to the corresponding categories, and are easier to be recognized than those in the raw stage (zooming in is recommended for better visualization).}
    \label{fig:heatmap}
\end{figure}

\subsection{Sensitivity Analysis}
As shown in Section~\ref{loss_function}, there is a hyper-parameter $\lambda$ in the final loss function to control the contribution of $\mathcal{L}_{\text {local}}$. We conduct a comparative experiment on CUB-$200$-$2011$ under $5$-way $1$-shot setting to study the effect of this factor across different backbones. From Fig.~\ref{fig:lambda}, when achieving the best performance, we have observed that the value of $\lambda$ on Conv-$64$ is relatively larger compared to the ResNet-$12$. The results show the fact that the local loss accounts for a larger proportion of the total loss for the shallower network and delays the appearance of performance saturation points. This is because, during the training phase, the shallow feature extractor has a weak feature expression ability and is unable to explore the inter-category relationship for FS-FGR. In all our experiments, we set the value of $\lambda$ to $0.1$ for the deeper network (\ie, ResNet-$12$ and ResNet-$18$), and $0.4$ for the shallow network (\ie, Conv-$64$) to achieve the final results.

\begin{figure}[t!]
	\centering
	\subfloat[Raw Stage]{
		\begin{minipage}{0.3\linewidth}
			\centering
			\includegraphics[width=1.0\textwidth]{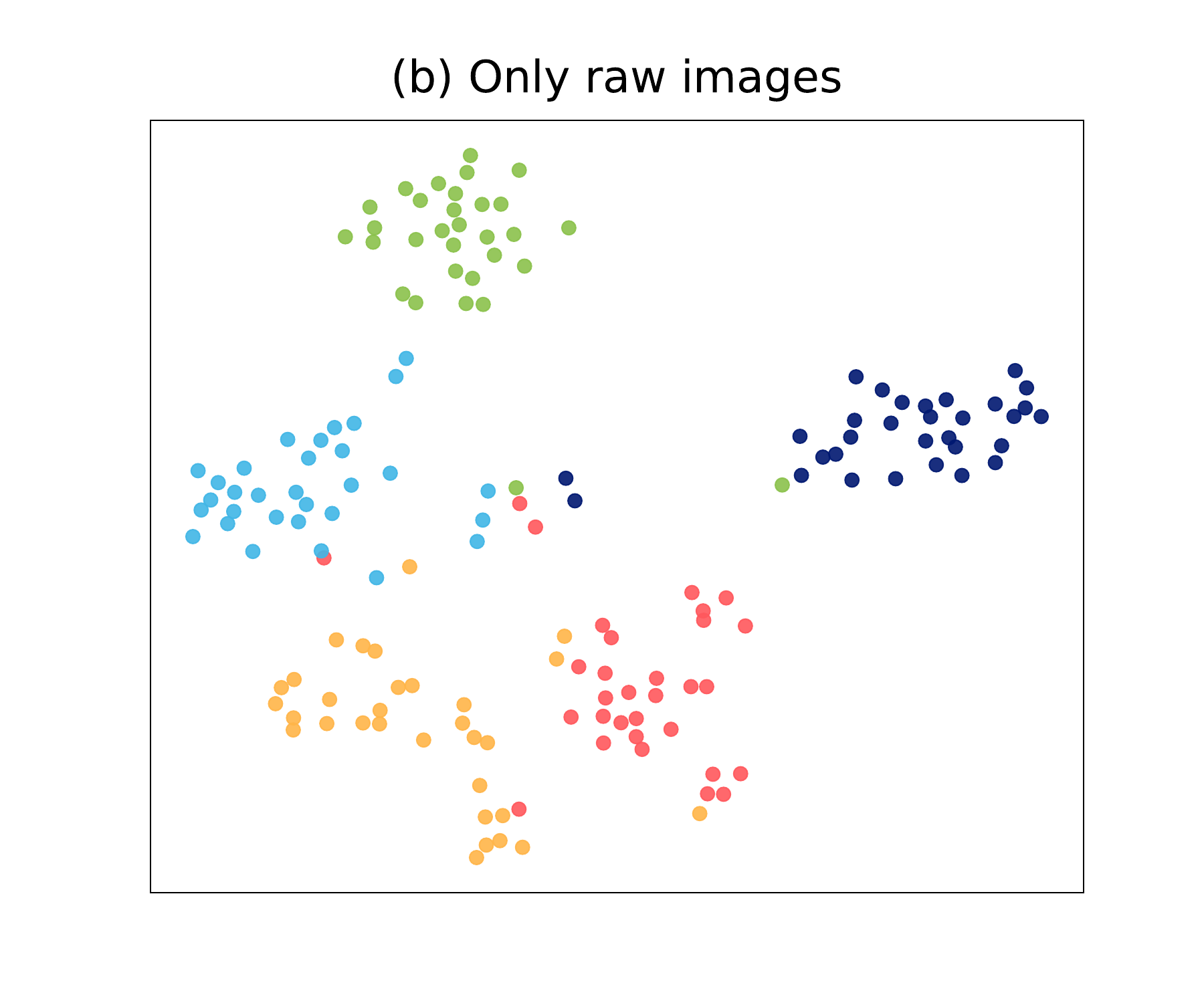}
		\end{minipage}
	}
	\subfloat[Refined Stage]{
		\begin{minipage}{0.3\linewidth}
			\centering
			\includegraphics[width=1.0\textwidth]{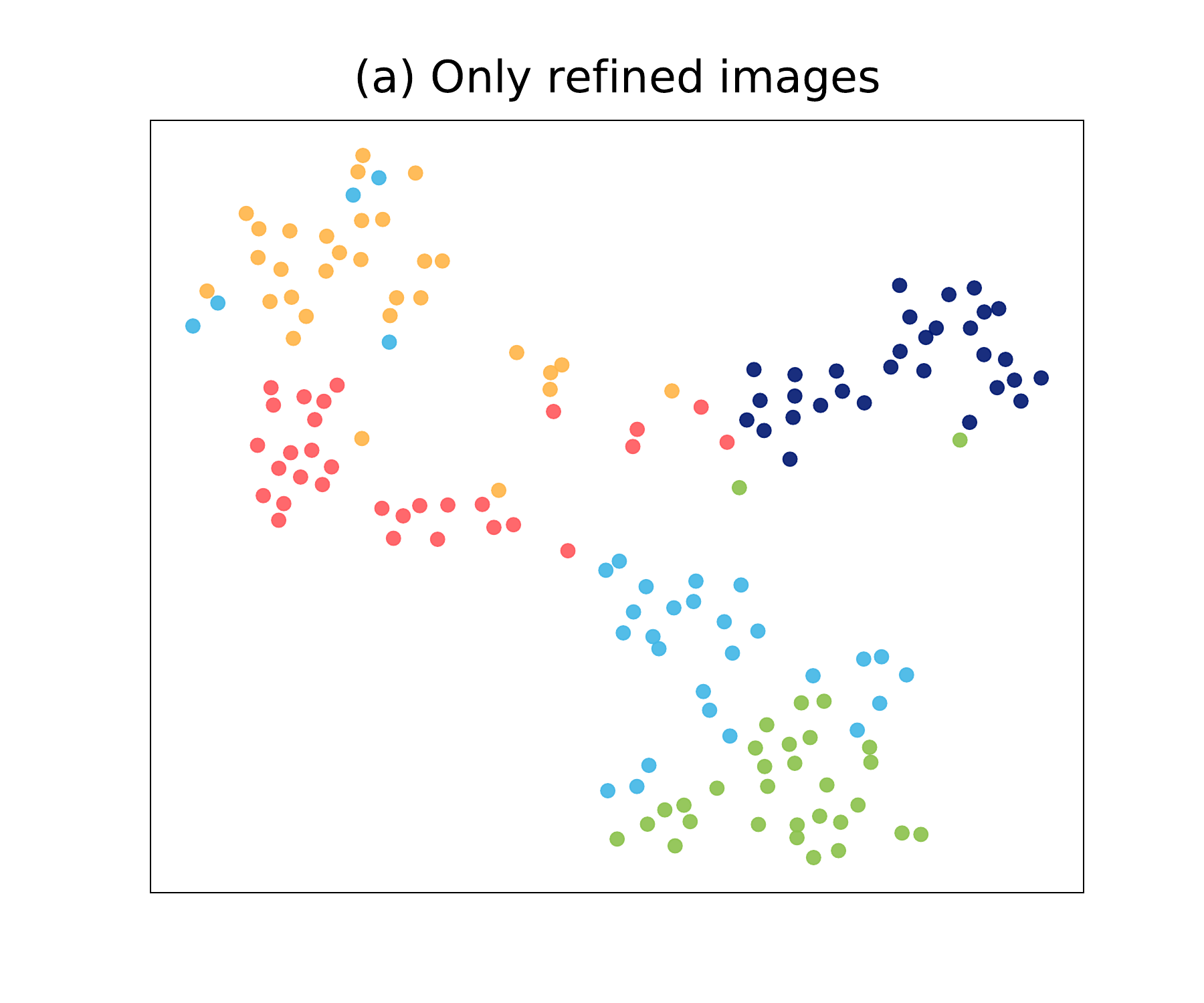}
		\end{minipage}
	}
	\subfloat[Raw + Refined Stages]{
		\begin{minipage}{0.3\linewidth}
			\centering
			\includegraphics[width=1.0\textwidth]{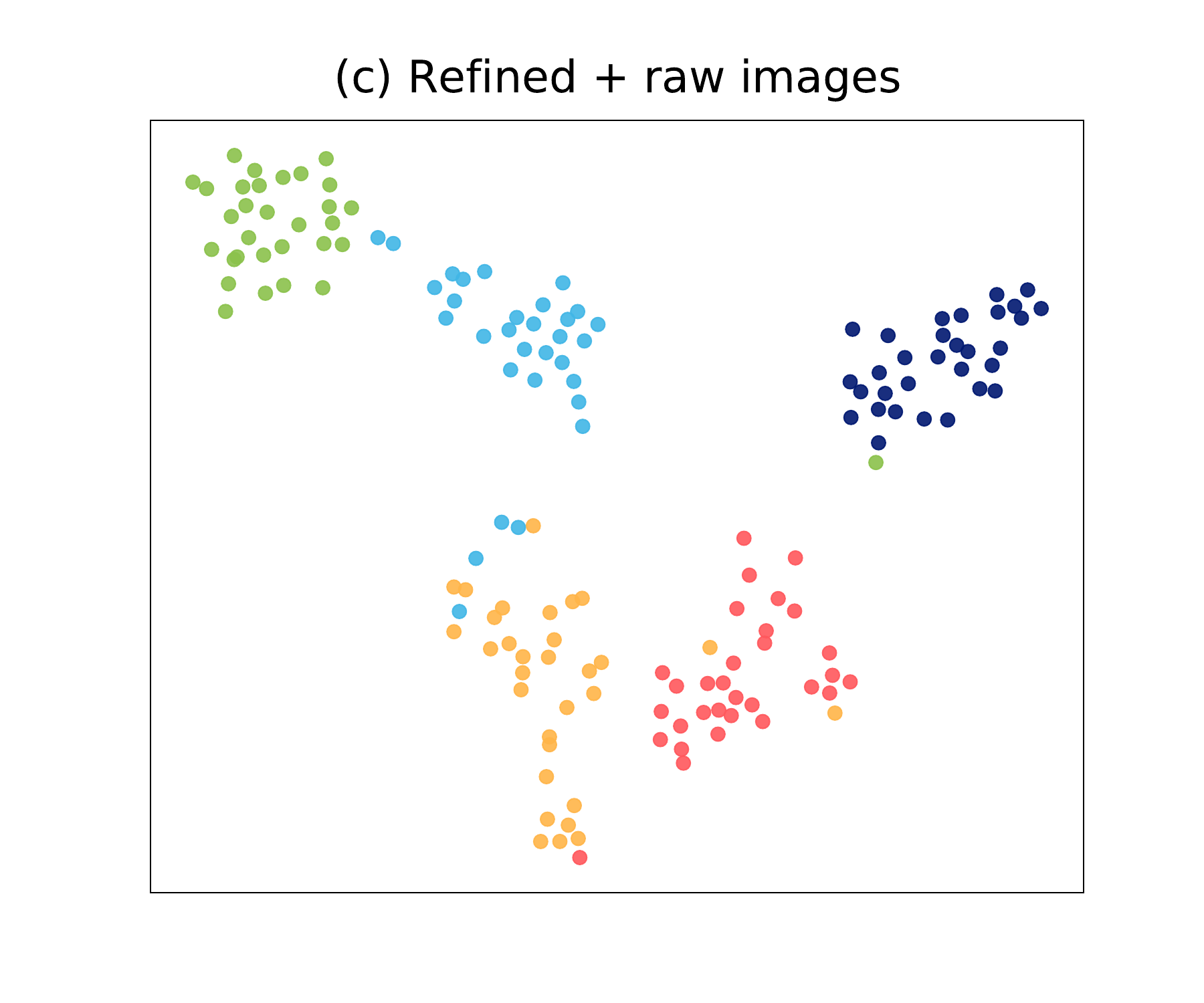}
		\end{minipage}
	}
	\caption{The t-SNE visualization of feature distribution in different stages on CUB-$200$-$2011$. We randomly select five classes (one for each color) with 30 query images in each class (one for each point). (a) Raw Stage: features are from only the raw stage. (b) Refined Stage: features are from only the refined stage. (c) Raw + Refined Stages: features are from both stages.}
	\label{fig:tnse}
\end{figure}

There are two weighting factors $\alpha$ and $\beta$ in global loss (\emph{i.e.~}Eq.~\ref{global_loss}) and local loss (\emph{i.e.~}Eq.~\ref{local_loss}) to control the contributions of raw image and refined image. The sensitivity analyses of two parameters according to $\alpha+\beta=1$ are performed on the CUB-200-2011 dataset and the results are presented in Table~\ref{tab:alpha_beta}. We can observe a significant gap in the $1$-shot performance between different settings of the weights. Meanwhile, although different backbones have different feature expression abilities and foreground predictions, the highest mean value of performance is achieved when the weights of the two factors both equal to $0.5$. The underlying reason could be that the raw stage and the refined stage are complementary to each other in improving the performance, which is also proved in Fig.~\ref{fig:tnse}. Thus, we simply select equal weights for the two weighting factors in the training and inference stage.

\subsection{Visualization Analysis}
In this section, we first visualize the results of background suppression on the CUB-$200$-$2011$. As presented in Fig.~\ref{fig:vis}(a)$-$(d), the proposed BAS module can cover the object region more completely and effectively remove the cluttered background, which enables the model to see more clearly with only image-level annotations. But as shown in Fig.~\ref{fig:vis}(e)$-$(g), we also find that the proposed BAS module performs poorer effectiveness for localizing smaller objects, 
especially the small object that is far away from the center of the image.

We gracefully argue that taking an amplified attended region without distraction from the background as input in the refine stage can distill more discriminative fine-grained characteristics in low-data scenarios. The insight behind it is that ``see closer, see better'' because the background noises can distract attention from the subtle differences. To verify this idea, we visualize the feature heatmaps of the two stages in Figure~\ref{fig:heatmap}, it is clear that our model focuses on more discriminative visual cues in the refined stage, which confirms the significance of our two-stage paradigm for FS-FGR.

In Fig.~\ref{fig:tnse}, we visualize the feature distribution in the embedding space by t-SNE~\cite{van2008visualizing} on CUB-200-2011 under 5-way 1-shot setting, where $30$ query images are randomly sampled from each category. It can be seen that the features obtained from our full model with both raw and refined stages have more compact and separable clusters than the other two models using only the raw stage or refined stage, which indicates that our method has good adaptability to the object's scale. Furthermore, the visualization of the model using only raw images is better than the one using only refined images. This further substantiates the correlation analysis in Section~\ref{ab} and indicates that the raw stage and the refined stage are complementary to each other in improving performance.

\begin{figure}[t!]
    \centering
    \includegraphics[scale=0.4]{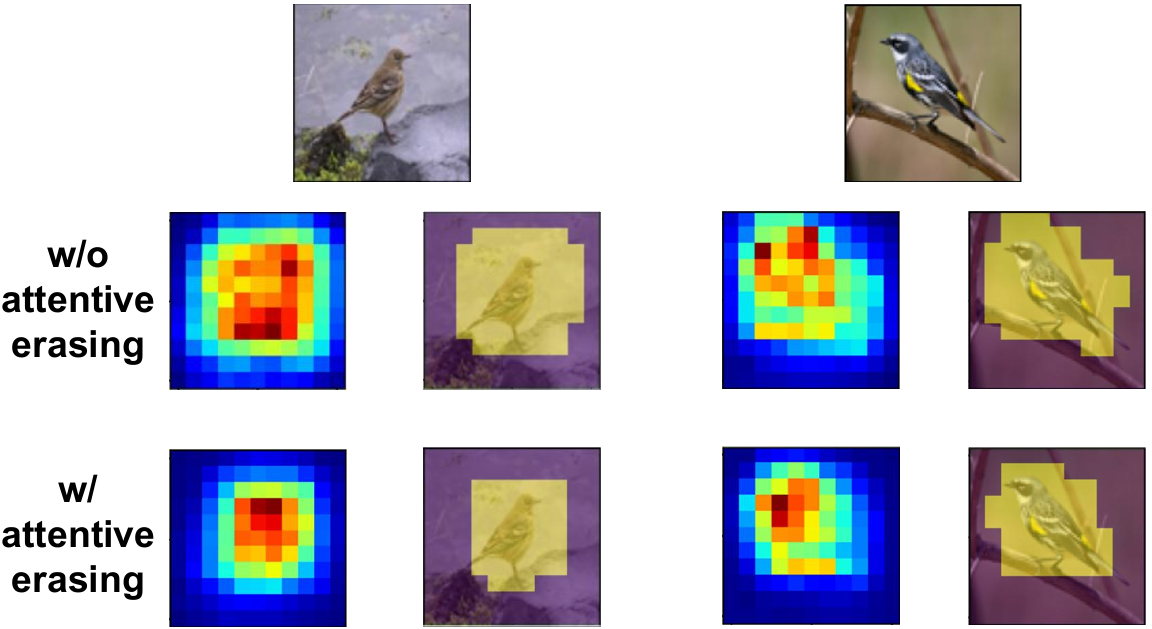}
    \vspace{-2mm}
    \caption{Visualization of the influence of attentive erasing strategy for foreground localization and background suppression.}
    \label{fig:cmp}
\end{figure}

Lastly, we investigate the effects of the attentive erasing strategy in foreground localization and background suppression. Fig.~\ref{fig:cmp} provides the visual results on CUB-200-2011, which proves the advantages of the proposed attentive erasing strategy. We observe that the proposed attentive erasing strategy facilitates the generation of foreground maps by indirectly increasing the foreground activation value, and the masked prediction map achieves better coverage of the localization results on the foreground object.
\vspace{-2mm}

\subsection{Discussion}

In this paper, we explore a two-stage weakly-supervised framework by introducing two parameter-free adjustments: \textit{background suppression} and \textit{foreground alignment}. Here, we also conduct some analyses to intuitively demonstrate the effectiveness and limitation of the proposed method.

To deal with the cluttered background, we propose a BAS module to disentangle the foreground object and background distraction in an image. As shown in Table~\ref{tab:diss}, `Ours(w/~BAS)' achieves significant and consistent improvement on both 1-shot and 5-shot tasks when compared to `Ours(w/o~BAS)'. But as seen in Fig.~\ref{fig:vis}, we note that our BAS is inconsistent for localizing objects of different sizes, especially poorer localization for small objects in the corner. Although the refined image generated by BAS covers the entire area of the foreground, it also contains some low-activated background regions when compared to the foreground region cropped by the ground-truth bounding box.

Going further, we attempt to achieve better localization of the foreground object by using BAS twice, \ie, Ours(w/~BAS$\times$2), which is devised to maximize the ratio of foreground activation and overall activation. It is a pity that the performance does not achieve further improvement as expected, which also indicates that the proposed BAS is insufficient to resist background distractions in some cases. The underlying reason could be that the limitation of feature map resolution causes imprecise localization, especially since the distribution between the foreground and background is unbalanced.

Lastly, we keep the raw image unchanged and replace the refined image with the image cropped by the ground-truth bounding box. As shown in Table~\ref{tab:diss}, `Ours(w/~BB.)' can be seen as an oracle and the upper bound of our method. Not surprisingly, `Ours(w/~BB.)' surpasses all existing methods by a large margin, which further demonstrates the potential of our proposed two-stage framework. Thus, how to overcome the limitation of the BAS module to achieve similar results with the ground-truth bounding box will be researched in future work.

\begin{table}[tb!]
	\centering
	\caption{ \label{tab:diss}The accuracy (\%) of $5$-way $1$-shot and $5$-shot tasks on CUB-200-2011 with various settings.}
	\vspace{-2mm}
	\resizebox{0.9\linewidth}{!}{ 
		\begin{tabular}{lccc}
			\toprule
			\specialrule{0em}{1pt}{1pt}
            \multirow{2}{*}{\bf Method} & \multirow{2}{*}{\bf Backbone} &\multicolumn{2}{c}{\bf CUB-200-2011} \\
			\cline{3-4}
			\specialrule{0em}{1pt}{1pt}
			& & {$5$-way $1$-shot} & {$5$-way $5$-shot}  \\
			\midrule
			\specialrule{0em}{1pt}{1pt}	
			Ours(w/o~BAS)  & ResNet-$12$& {79.75} {\scriptsize $\pm$ 0.48} & {89.44} {\scriptsize $\pm$ 0.28} \\
			\specialrule{0em}{1pt}{1pt}
			Ours(w/~BAS) & ResNet-$12$ & {82.27}  {\scriptsize $\pm$ 0.46} & {90.76} {\scriptsize $\pm$ 0.26} \\
			\specialrule{0em}{1pt}{1pt}	
			Ours(w/~BAS$\times$2)  & ResNet-$12$& {81.62} {\scriptsize $\pm$ 0.47} & {90.48} {\scriptsize $\pm$ 0.27} \\
			\specialrule{0em}{1pt}{1pt}	
			Ours(w/~BB.)  & ResNet-$12$& \bf{86.29} {\scriptsize $\pm$ 0.42} & \bf{93.23} {\scriptsize $\pm$ 0.22} \\
			\bottomrule
	\end{tabular}}	
	\vspace{-4mm}
\end{table}

\section{Conclusion} \label{Conclude}
In this paper, we show that background suppression and foreground alignment matter for few-shot fine-grained recognition (FS-FGR). Thus, we propose a novel two-stage weakly-supervised framework for FS-FGR, where the background suppression and foreground alignment are jointly learned in an end-to-end manner. In particular, a background activation suppression module is introduced in the raw stage to weaken background disturbance and enhance dominative foreground objects for the refined stage. In each stage, a foreground object alignment module is proposed to align the foreground object of support features concerning query features for addressing the problem of spatial misalignment. Finally, a local-to-local similarity metric is introduced to enable the model to capture subtle differences of confused sample pairs on two stages for the final decision. Extensive experiments on three fine-grained benchmarks demonstrate that our proposed model can be trained in an end-to-end manner with only the image-level label, and achieve state-of-the-art performance. What calls for special attention is that this is the first research to reveal the impact of background suppression and foreground alignment, and the first attempt to integrate both aspects into a standard pipeline for FS-FGR task. In addition, some designs in our designed generic framework include some heuristics, which could be further studied in detail.

\ifCLASSOPTIONcaptionsoff
  \newpage
\fi

{\small
\bibliographystyle{IEEEtran}
\bibliography{ref}
}

\end{document}